\def\cvprPaperID{4643}
\def\confName{ICCV}
\def\confYear{2023}

\def\paperTitle{Hierarchical Diffusion Autoencoders and Disentangled Image Manipulation}

\def\authorBlock{
    Zeyu Lu$^{1}$ \quad
    Chengyue Wu$^{2}$ \quad
    Xinyuan Chen$^{1}$ \quad
    Yaohui Wang$^{1}$ \quad
    Lei Bai$^{1}$ \quad
    Yu Qiao$^{1}$ \quad
    Xihui Liu$^{2}$ \quad
    \\
    \\
    $^{1}$ Shanghai AI Laboratory \qquad
    $^{2}$ The University of Hong Kong \qquad
}

\newif\ifreview 
\newif\ifarxiv \newcommand{\arxiv}{\arxivtrue}
\newif\ifcamera 
\newif\ifrebuttal 
\arxiv 

\pdfoutput=1
\documentclass[10pt,twocolumn,letterpaper]{article}
\ifreview \usepackage[review]{cvpr} \fi
\ifarxiv \usepackage[pagenumbers]{cvpr} \fi
\ifrebuttal \usepackage[rebuttal]{cvpr} \fi
\ifcamera \usepackage{cvpr} \fi

\usepackage{graphicx}
\usepackage{amsmath}
\usepackage{amssymb}
\usepackage{booktabs}
\usepackage{float}

\usepackage{times}
\usepackage{microtype}
\usepackage{epsfig}
\usepackage[table,xcdraw]{xcolor}
\usepackage{caption}
\usepackage{float}
\usepackage{placeins}
\usepackage{color, colortbl}
\usepackage{stfloats}
\usepackage{enumitem}
\usepackage{tabularx}
\usepackage{xstring}
\usepackage{multirow}
\usepackage{xspace}
\usepackage{url}
\usepackage{subcaption}
\usepackage{xcolor}
\usepackage[hang,flushmargin]{footmisc}
\ifcamera \usepackage[accsupp]{axessibility} \fi

% ...

\ifarxiv  \fi

\newcommand{\R}[1]{{%
    \textbf{%
        \ifstrequal{#1}{1}{\textcolor{red}{R#1}}{%
        \ifstrequal{#1}{2}{\textcolor{blue}{R#1}}{%
        \ifstrequal{#1}{3}{\textcolor{magenta}{R#1}}{%
        \ifstrequal{#1}{4}{\textcolor{teal}{R#1}}{%
                           \textcolor{cyan}{R#1}%
        }}}}%
    }%
}}

\usepackage{xr-hyper}

\makeatletter
\newcommand*{\addFileDependency}[1]{
  \typeout{(#1)}
  \@addtofilelist{#1}
  \IfFileExists{#1}{}{\typeout{No file #1.}}
}

\makeatother

\usepackage[pagebackref,breaklinks,colorlinks]{hyperref}
\usepackage[capitalize]{cleveref}
\crefname{section}{Sec.}{Secs.}
\crefname{table}{Table}{Tables}
\crefname{figure}{Fig.}{Figs.}

\begin{document}
\title{\paperTitle}
\author{\authorBlock}


\twocolumn[{
    \renewcommand\twocolumn[1][]{#1}
    \maketitle 
    \begin{center}
\vspace{-0.8cm}
    \centering
    \includegraphics[width=0.9\textwidth]{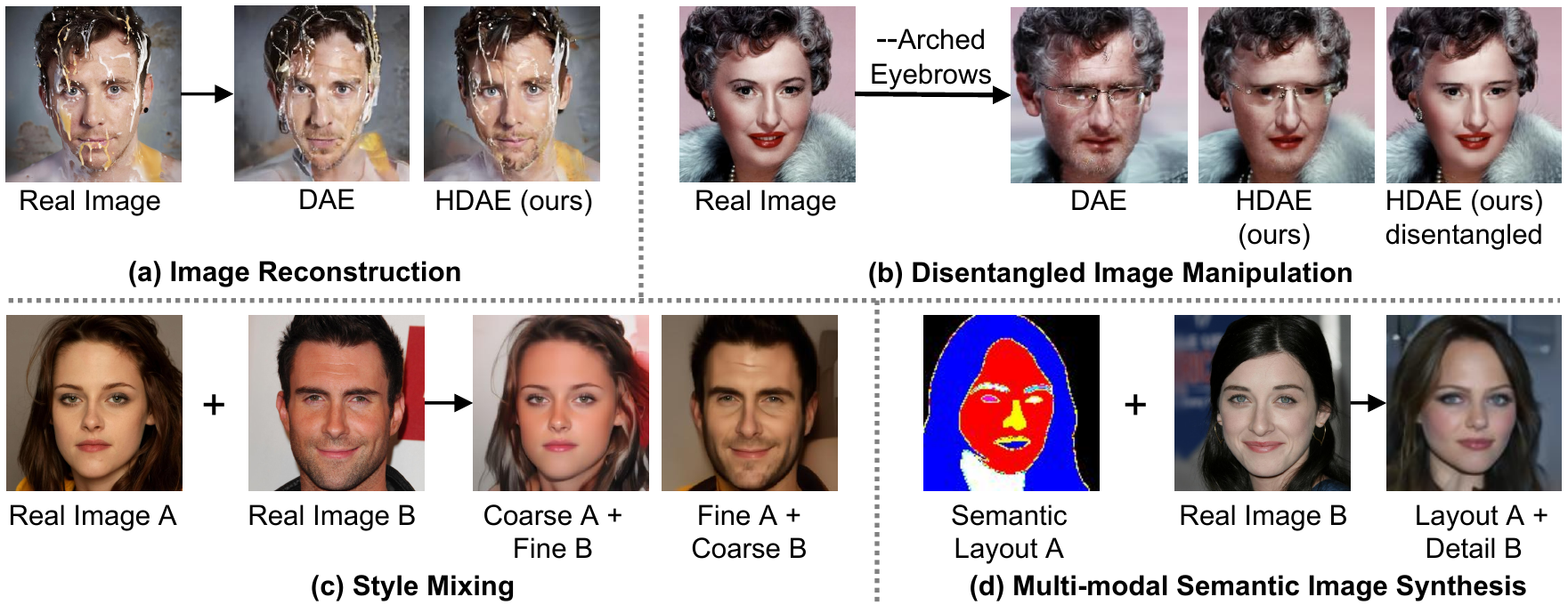}
    \vspace{-0.1cm}
    \captionof{figure}{
        \textbf{Applications of Hierarchical Diffusion Autoencoders (HDAE).}
        (a) Near-perfect image reconstruction. 
        (b) Disentangled image manipulation. Our approach disentangles ``arched eyebrows'' with other related attributes such as ``female'' and ``eyeglasses''. 
        (c) Style mixing with different levels of features from different images. 
        (d) Multi-modal semantic image synthesis with a layout image and a real image providing information on style and details.
    }
    \vspace{-0.1cm}
    \label{teaser}  
    
\end{center}

}]

\maketitle 
\begin{abstract}
Diffusion models have attained impressive visual quality for image synthesis.
However, how to interpret and manipulate the latent space of diffusion models has not been extensively explored.
Prior work diffusion autoencoders encode the semantic representations into a semantic latent code, which fails to reflect the rich information of details and the intrinsic feature hierarchy.
To mitigate those limitations, we propose Hierarchical Diffusion Autoencoders (HDAE) that exploit the fine-grained-to-abstract and low-level-to-high-level feature hierarchy for the latent space of diffusion models.
The hierarchical latent space of HDAE inherently encodes different abstract levels of semantics and provides more comprehensive semantic representations.
In addition, we propose a truncated-feature-based approach for disentangled image manipulation.
We demonstrate the effectiveness of our proposed approach with extensive experiments and applications on image reconstruction, style mixing, controllable interpolation, detail-preserving and disentangled image manipulation, and multi-modal semantic image synthesis.


\end{abstract}
\vspace{-10pt}
\section{Introduction}

%

Diffusion models~\cite{DDPM,DULNT} have demonstrated impressive image generation quality and
achieved remarkable success in various applications, such as text-to-image generation~\cite{DALLE2,GLIDE,Rombach2022latent}, image editing~\cite{p2p, DAE, Dreambooth}, and inpainting~\cite{repaint,NTIRE}.

A semantically meaningful, editable, and decodable latent space is of particular importance in interpreting generative models as well as applications such as image editing.
There have been various works on learning visual representations and manipulating the latent space of GANs~\cite{Interfacegan,SeFa,GANspace}.
However, the properties of the representations learned by diffusion models are underexplored.
Preechakul~\etal proposed Diffusion Autoencoders (DAE)~\cite{DAE}, which leverages a learnable encoder to discover high-level semantic representations and a diffusion model to encode the stochastic variations and decode images from the semantic latent code and the stochastic latent code.

However, the semantic latent code of diffusion autoencoders is simply represented by a holistic feature vector predicted by the final layer features of the semantic encoder. 
Such a latent space is insufficient to encode rich information, particularly low-level and mid-level features. 
In practice, we observe that the latent representation omits the fine-grained features (\eg, background and low-level details), leading to imperfect image reconstruction and editing results.
Moreover, it cannot reflect the coarse-to-fine and low-level-to-high-level hierarchy of feature representations.

\begin{figure*}[!h]
    \centering
    \includegraphics[width=0.95\linewidth]{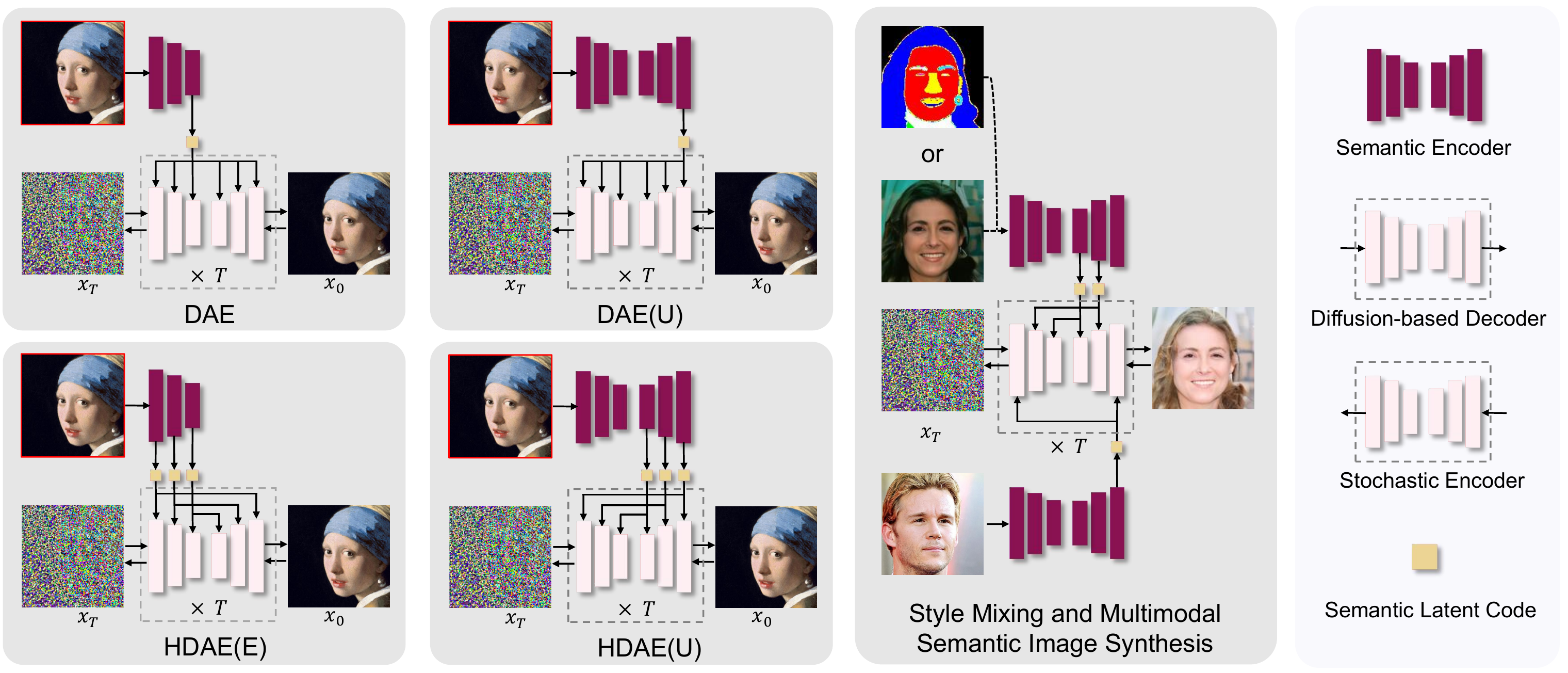}
    \vspace{-0.1cm}
    \caption{
        \textbf{Overview of different network structures.}
        In general, diffusion autoencoders apply the semantic encoder to encode the semantic latent code, the stochastic encoder (\ie, the DDIM forward process) to encode the stochastic latent code, and the diffusion-based decoder (\ie, the DDIM reverse process) to generate images based on the latent codes.
        DAE~\cite{DAE} and DAE(U) are non-hierarchical diffusion autoencoders.
        HDAE(E) and HDAE(U) are different variants of our proposed hierarchical diffusion autoencoders. In addition, we show style mixing and multimodal image synthesis with HDAE, where the low-level latent code and high-level latent code are from different images.
    }
    \vspace{-0.2cm}
    \label{overview}
\end{figure*}

To mitigate those problems, we design the \textbf{Hierarchical Diffusion Autoencoders} (HDAE) that exploits the coarse-to-fine and low-level-to-high-level feature hierarchy of the semantic encoder and the diffusion-based decoder for comprehensive and hierarchical representations.
Our design of the hierarchical latent space is motivated by the observation that feature maps at different scales correspond to different abstraction levels of features.
The high-resolution feature maps contain low-level features (\eg, color, texture, and details) and the low-resolution feature maps contain high-level features (\eg, structure, layout, abstract attributes).
In particular, we extract different levels of features from the semantic encoder and use them to predict the semantic latent codes for the corresponding feature levels of the diffusion-based decoder.
We extensively investigate different design choices of HDAE, as demonstrated in Fig.~\ref{overview}, and found that HDAE with U-Net encoder, denoted as HDAE(U), achieves the best performance.
Our HDAE(U) converges faster and better than DAE~\cite{DAE}.
Image reconstruction and semantic manipulation experiments demonstrate that the latent representations of HDAE are richer and more comprehensive than DAE.
Moreover, the hierarchical representations enable more applications than DAE such as style mixing, controlled interpolation, and multimodal semantic image synthesis.

The latent semantic representations provide a natural way to edit images by moving the latent features toward the specific direction of a semantic attribute.
However,  different attributes are entangled within the latent space, making it difficult to find a single direction to manipulate a specific attribute without affecting other attributes.
To further improve the disentanglement of attributes for image manipulation,
we propose to conduct manipulation with truncated features, based on the observation that the majority of feature channels are with low values, and those feature channels are a critical cause of entanglement.
Experiments demonstrate that the truncated features facilitate disentangled attribute manipulation of face images, \eg, we can disentangle ``old'' from ``wearing eyeglasses'' to edit a face image towards older without adding eyeglasses to the face.

In summary, our contributions are as follows.
(1) We propose Hierarchical Diffusion Autoencoders which exploit the coarse-to-fine and low-level-to-high-level features to obtain a comprehensive and hierarchical latent space for diffusion models.
(2) We propose a truncated-feature-based method for disentangled image manipulation with hierarchical representations.
(3) We demonstrate the effectiveness of our approach by image reconstruction, style mixing, controlled interpolation, multimodal semantic image synthesis, and disentangled image manipulation, shown in Fig.~\ref{teaser}.

\label{sec:intro}

\section{Related Work}

\noindent\textbf{Diffusion models.}
Diffusion models have shown great capability in image synthesis.
Song ~\etal~\cite{NCSN} proposed score-based generative models as a way of modeling a data distribution using its gradients.
Ho~\etal~\cite{DDPM} proposed denoising diffusion probabilistic models (DDPMs) which achieved high sample quality based on the score-based generative models~\cite{NCSN} and the diffusion models~\cite{DULNT}.
Inspired by the progress, many works improved the sampling speed~\cite{DDIM,AnalyticDPM}, sampling quality~\cite{IDDPM,ImproveNCSN}, and conditional synthesis~\cite{ADM}.
Diffusion models have also shown wide applications in text-to-image generation~\cite{DALLE2,GLIDE,imagen,Rombach2022latent}, image translation~\cite{Palette, SDEdit, PITI}, image editing~\cite{p2p, DAE, Dreambooth, MCF}, image inpainting~\cite{repaint,NTIRE}, speech generation~\cite{WaveGrad} and text generation~\cite{DLMICTG,SDDM}.

\noindent\textbf{Latent space of generative models.}
Researchers have made attempts to interpret and manipulate the latent space of generative models.
The StyleGAN~\cite{Stylegan} generator maps the random noise vector to a semantically meaningful latent space, inspiring various follow-up works exploring the controllability and interpretability of the latent space of GANs~\cite{StyleSpace, BDInvert, PULSE, Stylegan, DCGAN, Toc2021e4e, roich2021pivotal}.
Various works explored learning based~\cite{GHFeat,PSP,HFGI}, optimization based~\cite{Image2StyleGAN, Image2StyleGAN++, StyleSpace, BDInvert}, or hybrid~\cite{HyperStyle,GVMNIM} approaches for GAN inversion, aiming to encode an image to the latent space of GANs.
GAN inversion enables diverse image editing methods by manipulating the latent code.
Shen~\etal~\cite{Interfacegan} proposed InterfaceGAN which adopts an SVM to find the semantic directions for attribute manipulation.
H{\"{a}}rk{\"{o}}nen~\etal~\cite{GANspace} proposed GANspace which performed PCA on early feature layers.
Some methods~\cite{UDDMG,EYE,LatentCLR} found distinguishable directions based on mutual information.
Recently some works~\cite{Styleclip, diffusionclip,clippca,StyleMC} explored image editing guided by CLIP~\cite{CLIP} model.
In particular, Patashnik~\etal~\cite{Styleclip} proposed StyleClip which used the pre-trained CLIP~\cite{CLIP} as the loss supervision to match the manipulated results with the text condition.
Our work explores the hierarchical latent space of diffusion models.


\noindent\textbf{Latent space of diffusion models.}
Despite the various work studying the latent space of GANs, the latent space of diffusion models lack semantic meaning and cannot be easily applied for semantic manipulation.
Latent diffusion~\cite{Rombach2022latent} applies diffusion model in the latent space of images instead of the original image space.
DiffusionCLIP~\cite{diffusionclip} conducts image editing by DDIM inversion and model finetuning guided by CLIP. However, the latent space of diffusion models cannot be directly manipulated for image editing.
Diffusion autoencoders~\cite{DAE} adopts a semantic encoder to obtain a meaningful and decodable latent space for diffusion models.
However, such latent space lacks find details and causes attribute entanglement.
We propose the Hierarchical Diffusion Autoencoders which provides a comprehensive and hierarchical latent space for diffusion models.

\label{sec:related}

\section{Methodology}


\subsection{Preliminaries}\label{subsec:pre}

\noindent\textbf{Diffusion probabilistic models.}
Denoising diffusion probabilistic models (DPMs)~\cite{DDPM} is a class of generative models. The forward process defines a Markov chain gradually adding Gaussian noise to an image $x_0$, generating a sequence of images $x_0, x_1, \cdots, x_T$.
The reverse process iteratively removes noise from the noisy image $x_t$ by sampling from $p(x_{t-1}|x_t)$.
The noise $\epsilon_{\theta}(x_t, t)$ is predicted by a U-Net which takes the noisy image $x_t$ and timestep $t$ as input.
The model is trained with the $L_2$ loss between the predicted noise and the actual noise $\lVert \epsilon-\epsilon_\theta(x_t,t)\rVert_2^2$



Song~\etal~\cite{DDIM} proposed Denoising Diffusion Implicit Model (DDIM) with a deterministic forward process. 
By matching the marginal distribution of DDPM, it shares the same training objective and solution with DDPM.
We can run the DDIM generation process backward deterministically to obtain the noise map $x_T$, which represents the stochastic latent codes of the image $x_0$.

\noindent\textbf{Diffusion autoencoders.}
In pursuit of a meaningful latent space, Preechakul~\etal~\cite{DAE} proposed Diffusion Autoencoders.
They apply a convolutional neural network as the semantic encoder to encode images into a semantic vector $z_s=\text{Enc}_\phi(x_0)$ and apply the DDIM forward process as a stochastic encoder that encodes the image $x_0\in \mathbb{R}^{H \times W \times 3}$ to a stochastic variant $x_T\in \mathbb{R}^{H \times W \times 3}$. The DDIM reverse process acts as the decoder which models $p_\theta(x_{t-1}|x_t, z_s)$ and iteratively generates $x_0$ given the semantic latent code $z_s$ and the stochastic latent code $x_T$.
The DDIM adopts a U-Net structure with shared parameters for each timestep, and the U-Net is conditioned by the semantic code $z_s$ and timestep $t$ by adaptive group normalization layers (AdaGN).
The semantic latent vector $z_s\in \mathbb{R}^{512}$ captures the meaningful and decodable representations that can be used for image reconstruction and manipulation.


\label{sec:background}


\subsection{Hierarchical Diffusion Autoencoders}\label{subsec:hierarchical}

The semantic latent code in diffusion autoencoders is a 512-dimensional feature vector.
Such representations bring two limitations. 
Firstly, the single feature vector from the final layer of the semantic encoder omits the low-level and mid-level features, making it not sufficient to encode the comprehensive semantic information in the images.
Empirically, we observe that the images reconstruction and manipulation results with diffusion autoencoders suffer from insufficient details.
Secondly, the holistic feature representation ignores the intrinsic fine-grained-to-abstract and low-level-to-high-level hierarchy of visual features.

\noindent\textbf{Design space of latent representations and network architectures.}
To address those issues, we propose hierarchical diffusion autoencoders which explore the hierarchical semantic latent space of diffusion autoencoders.
In particular, we keep the architecture of the diffusion-based decoder and explore the design space of the semantic encoder and the latent space, as shown in Fig.~\ref{overview}.
\begin{itemize}
    \item[--]\textit{DAE, \ie, diffusion autoencoders.} The Na\"ive diffusion autoencoders~\cite{DAE} leverage a na\"ive CNN-based semantic encoder, where the semantic code is extracted from the last layer of the semantic encoder, followed by global average pooling and fully-connected layers.
    \item[--]\textit{DAE(U), \ie, diffusion autoencoders with U-Net semantic encoder.} We replace the na\"ive encoder in DAE with a U-Net encoder, denoted by DAE(U).
With the skip connections and downsampling-upsampling design, the last layer of the U-Net might be able to capture both low-level and high-level features.
The spatial feature map from the last layer of the U-Net is mapped into a 512-dimensional feature vector $z_s$.
    \item[--]\textit{HDAE(E), \ie, hierarchical diffusion autoencoders with na\"ive semantic encoder.} To exploit the feature hierarchy of the diffusion autoencoders,
we extract different semantic levels of feature maps from the semantic encoder to predict the hierarchical semantic latent codes $z_s^1, z_s^2, \cdots, z_s^L$.
The different levels of semantic features are fed into the corresponding levels of the diffusion-based decoder.
    \item[--]\textit{HDAE(U), \ie, hierarchical diffusion autoencoders with U-Net semantic encoder.} HDAE(U) also adopts the hierarchical latent space design but leverages a semantic encoder with the U-Net structure.
\end{itemize}

Experiments demonstrate that the hierarchical structures of HDAE(E) and HDAE(U) provide richer semantic representations and more efficient training, and that HDAE(U) is the best-performed architecture design. 

\noindent\textbf{Advantages and applications of HDAE.}
Firstly, while the latent space of DAE lacks low-level details, the hierarchical latent space of HDAE encodes comprehensive fine-grained-to-abstract and low-level-to-high-level features, leading to more accurate and detail-preserving image reconstruction and manipulation results. 
Secondly, the feature hierarchy naturally enables applications such as style mixing (illustrated in Fig.~\ref{overview}), multimodal semantic image synthesis, and controllable image interpolation, as demonstrated in Sec.~\ref{sec:exp}.

\begin{figure}[!t]
    \centering
    \includegraphics[width=0.98\linewidth]{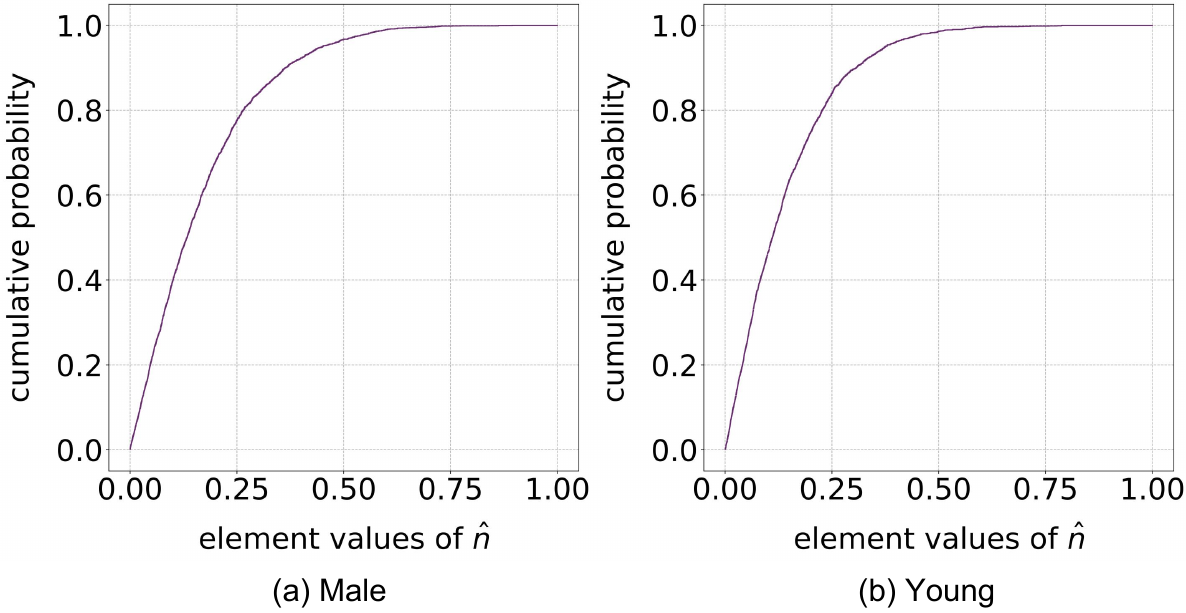}
    \caption{
        \textbf{The empirical cumulative distribution function of the element values in the normalized classifier weights $\hat{n}$.}
        Most elements of $\hat{n}$ are of low values, and only a few are of high values. 
    }
    \vspace{-0.28cm}
    \label{directionbarmap}
\end{figure}

\subsection{Disentangled Image Manipulation with Truncated Features}\label{subsec:disentangled}

With a linear classifier trained with the semantic vectors, diffusion autoencoders~\cite{DAE} can be applied for image manipulation.
We can edit an image by moving the semantic vector towards the direction $n$, which is obtained from the weights of the linear classifier for the target attribute.

A critical issue for image editing is the entanglement of features in the latent space.
For example, in face editing, ``old'' is often entangled with ``wearing glasses'', and ``arched eyebrows'' is often entangled with ``female''.
By analyzing the distribution of the classifier direction $n$, we propose an approach to disentangle attributes by adjusting $n$.

In our HDAE models with $L$ hierarchical layers, we concatenate the 512-dimensional semantic codes from each of the $L$ layers into a single vector and derive the classifier direction $n\in\mathbb{R}^{512\times L}$.
We derive the normalized classifier weights $\hat{n}$ as follows:
\begin{align}
    \hat{n}_i = \frac{|n_i| - \underset{i}{\operatorname{min}}(|n_i|)} {\underset{i}{\operatorname{max}}(|n_i|) - \underset{i}{\operatorname{min}}(|n_i|)}
\end{align}

\begin{figure}[!t]
    \centering
    \includegraphics[width=1\linewidth]{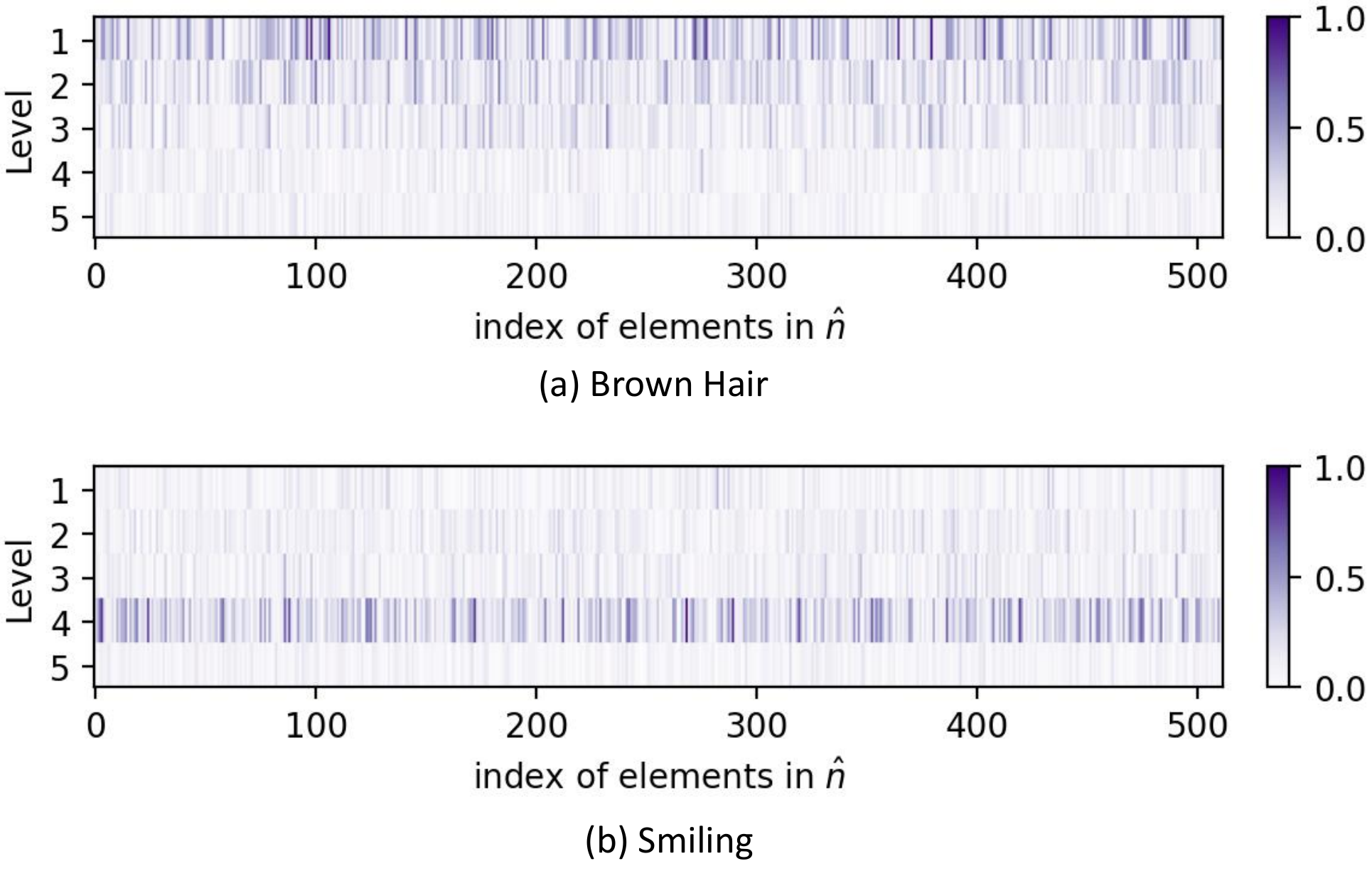}
    \caption{
        \textbf{The values of the $5\times512$-dimensional $\hat{n}$, visualized by levels.}
        (1) Most values are of low values and truncating those values lead to better disentanglement. (2) Feature hierarchy is shown.
    }
    \vspace{-0.2cm}
    \label{clsheatmap}
\end{figure}

As shown by the empirical cumulative distribution function and the visualization of the values of $\hat{n}$ in Fig.~\ref{directionbarmap} and Fig.~\ref{clsheatmap}, most values are relatively low, and only a few values are high.
We hypothesize that the few high-value elements indicate the dominant direction of an attribute classifier, while the majority, low-value elements are noisy and may lead to attribute entanglement.
In particular, we denote the set of the top-k largest values of $\hat{n}$ as $\text{top-k}(\hat{n})$, and truncate the $n$ accordingly as follows.
\begin{equation}
    n^\prime_i=\left\{\begin{array}{l}
         n_i, \text{ if } \hat{n}_i\in  \text{top-k}(\hat{n})  \\
         0, \text{   else} 
    \end{array}
    \right.
\end{equation}

Our experiments in Fig.~\ref{topkmanipulate} validate that the truncated $n^\prime$ leads to better-disentangled image manipulation.


\label{sec:method}

\section{Experiments}\label{sec:exp}
\subsection{Experimental settings}

Following Diffusion Autoencoders~\cite{DAE}, we train the hierarchical diffusion autoencoders on FFHQ~\cite{Stylegan} dataset and train the attribute classifiers on CelebA-HQ~\cite{ProGAN} dataset.


\begin{figure}[!t]
    \centering
    \includegraphics[width=0.95\linewidth]{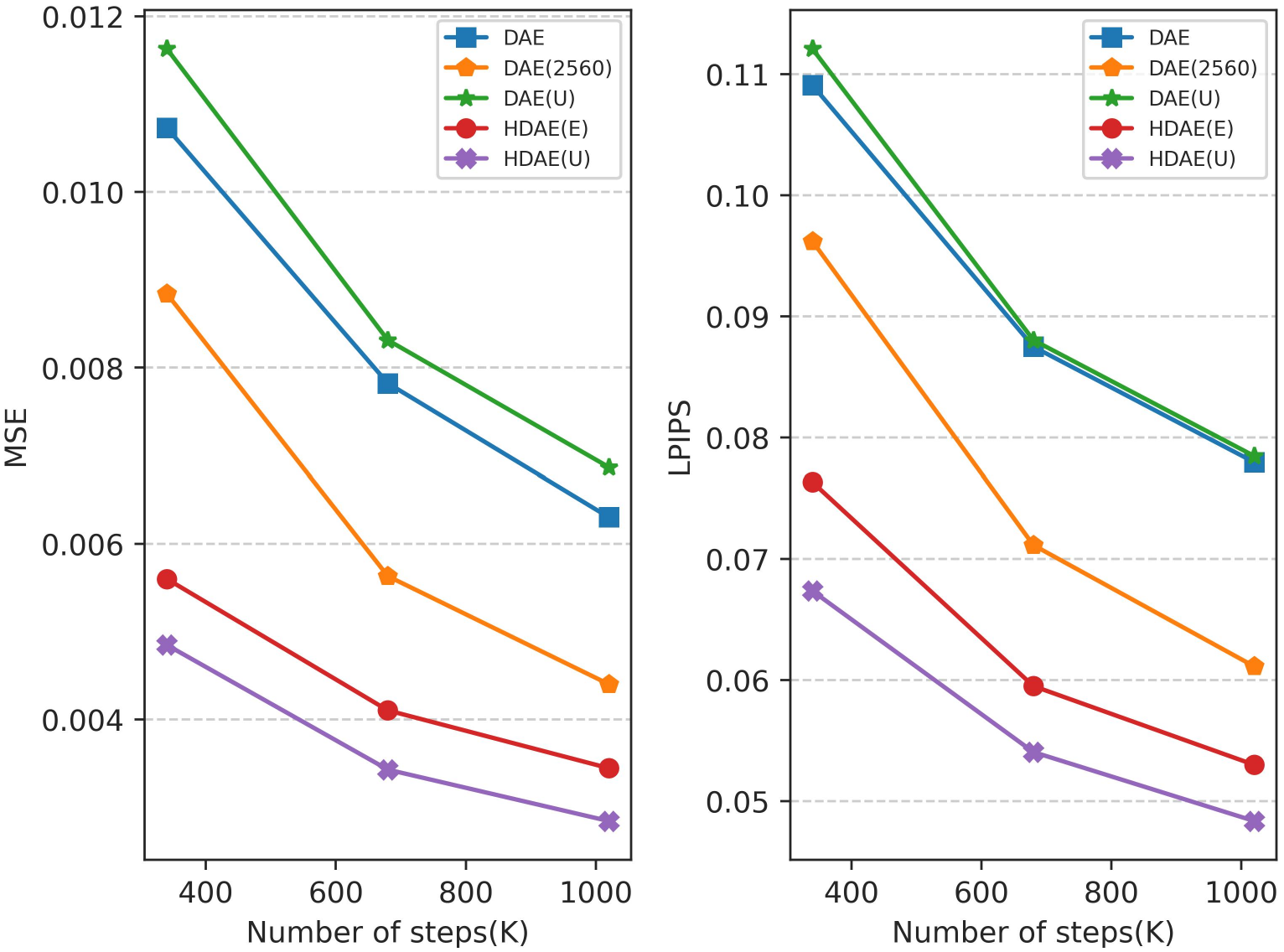}
    \caption{
        \textbf{Ablation study of different architecture designs for image reconstruction on FFHQ dataset.}
        MSE and LPIPS are evaluated on the test set.
        Both variants of HDAE outperform DAE baselines.
        HDAE(U) achieves the best image reconstruction performance.
        We adopt HDAE(U) as our best model in other experiments.
    }
    \vspace{-0.4cm}
    \label{ablationplot}
\end{figure}

The ablation study in Sec.~\ref{sec:abl} is trained on $128 \times 128$ images and other results are derived from models trained on $256 \times 256$ images.
More implementation details and experimental settings can be found in the appendix.

\subsection{Ablation Study}\label{sec:abl}
We conduct ablation studies to test the effectiveness of the design choices of diffusion autoencoders and hierarchical diffusion autoencoders demonstrated in Fig.~\ref{overview}.
We split the FFHQ dataset into 65,000 images for training and 5,000 for testing.
For a fair comparison, we add another baseline, denoted as DAE(2560), where the semantic code dimension of DAE is expanded from 512 to 2560, sharing the same size as the semantic code dimension ($512\times 5$) of our HDAE models.
We report the image reconstruction quality on the test set, evaluated by pixel-wise MSE and the perceptual quality, in Fig.~\ref{ablationplot}.
We draw the following conclusions:
\begin{itemize}
    \item[--]\textit{The hierarchical diffusion autoencoders perform much significantly better than non-hierarchical ones.}
As illustrated in Fig.~\ref{ablationplot}, both variants of hierarchical diffusion autoencoders, HDAE(E) and HDAE(U), perform significantly better and converge much faster than DAE, DAE(2560), and DAE(U). The image reconstruction performances of HDAE(E) and HDAE(U) at 340K steps are better than the performances of DAE and DAE(U) at 1,020K steps.
    \item[--]\textit{The U-Net structure for the semantic encoder has a negative effect on DAE, but benefits HDAE.} As demonstrated in Fig.~\ref{ablationplot}, DAE(U) performs worse than DAE, but HDAE(U) performs better than HDAE(E).
    \item[--]\textit{HDAE(U) is the best-performing model.} Therefore, we adopt HDAE(U) in the following experiments.
\end{itemize}




\begin{table}[t]
\resizebox{1\linewidth}{!}{
    \begin{tabular}{l|c|ccc}
    \toprule[1.2pt]
    \textbf{Model} & \textbf{Setting}  & \textbf{SSIM$\uparrow$}                & \textbf{LPIPS$\downarrow$}               & \textbf{MSE$\downarrow$}                   \\ \midrule[1.2pt]
    StyleGAN2\begin{math}(\mathcal{W})\end{math}~\cite{Stylegan2} &  -   & 0.677                        & 0.168                        & 0.016                          \\
    StyleGAN2\begin{math}(\mathcal{W+})\end{math}~\cite{Stylegan2} &  -  & 0.827                        & 0.114                        & 0.006                          \\
    VQ-GAN~\cite{VQGAN}&      -      & 0.782                        & 0.109                        & 3.61e-3                        \\
    VQ-VAE2~\cite{VQVAE2}&     -      & 0.947                        & 0.012                        & 4.87e-4                        \\
    HFGI~\cite{HFGI} & - & 0.877 & 0.127 & 0.0617 \\
    DDIM~\cite{DDIM} & T=100, $128^2$             & 0.917                        & 0.063                        & 0.002                          \\ \midrule[1.2pt]
    DAE~\cite{DAE} & T=100, $128^2$, random $x_T$   & 0.677                        & 0.073                        & 0.007                          \\
    \textbf{HDAE(U) (ours)} & T=100, $128^2$, random $x_T$ &  0.793 &  0.038 &  2.96e-3 \\ \midrule[1.2pt]
    DAE~\cite{DAE} & T=100, $128^2$, encoded $x_T$   & 0.991                        & 0.011                        & 6.07e-5                        \\
    \textbf{HDAE(U) (ours)} & T=100, $128^2$, encoded $x_T$ & \textbf{0.993}               & \textbf{0.009}               & \textbf{5.01e-5}               \\ 
    \bottomrule[1.2pt]
    \end{tabular}
}
\caption{
    \textbf{Image reconstruction evaluation of models trained on FFHQ~\cite{Stylegan} and tested on CelebA-HQ~\cite{ProGAN}.}
    HDAE(U) outperforms DAE with random stochastic code $x_T$, indicating that HDAE(U) encodes richer details than DAE.
    HDAE(U) with encoded $x_T$ achieves state-of-the-art, near-perfect reconstruction.
}
\vspace{-0.3cm}
\label{imgrectab}
\end{table}
\begin{table}[t]
\resizebox{1\linewidth}{!}{
\begin{tabular}{l|ccccc}
\toprule[1.2pt]
\textbf{Task}     & \textbf{HDAE(U)} & \textbf{DAE} & \textbf{HDAE(U)+TF} & \textbf{DAE+TF} & \textbf{Similar} \\ \midrule[1.2pt]
Image Reconstruction & 82.5\%   & 11.7\% & -& -& 5.8\%   \\ 
Image Manipulation  & 65.7\%   & 3.6\% & -& -& 30.7\%  \\
Disentangled Image Manipulation  & 22.8\% & 1\% & 76.2\% & 0\% & -  \\
\bottomrule[1.2pt]
\end{tabular}
}
\caption{
    \textbf{Human perceptual evaluation on image reconstruction, image manipulation and disentangled image manipulation.} Since DAE and HDAE(U) achieve near-perfect reconstruction for naked human eyes, we conduct the human perceptual evaluation on image reconstruction with random stochastic code $x_T$. “+TF” denotes image manipulation with truncated features.
}
\vspace{-0.3cm}
\label{userstudytab}
\end{table}
\begin{figure}[!t]
    \centering
    \includegraphics[width=1\linewidth]{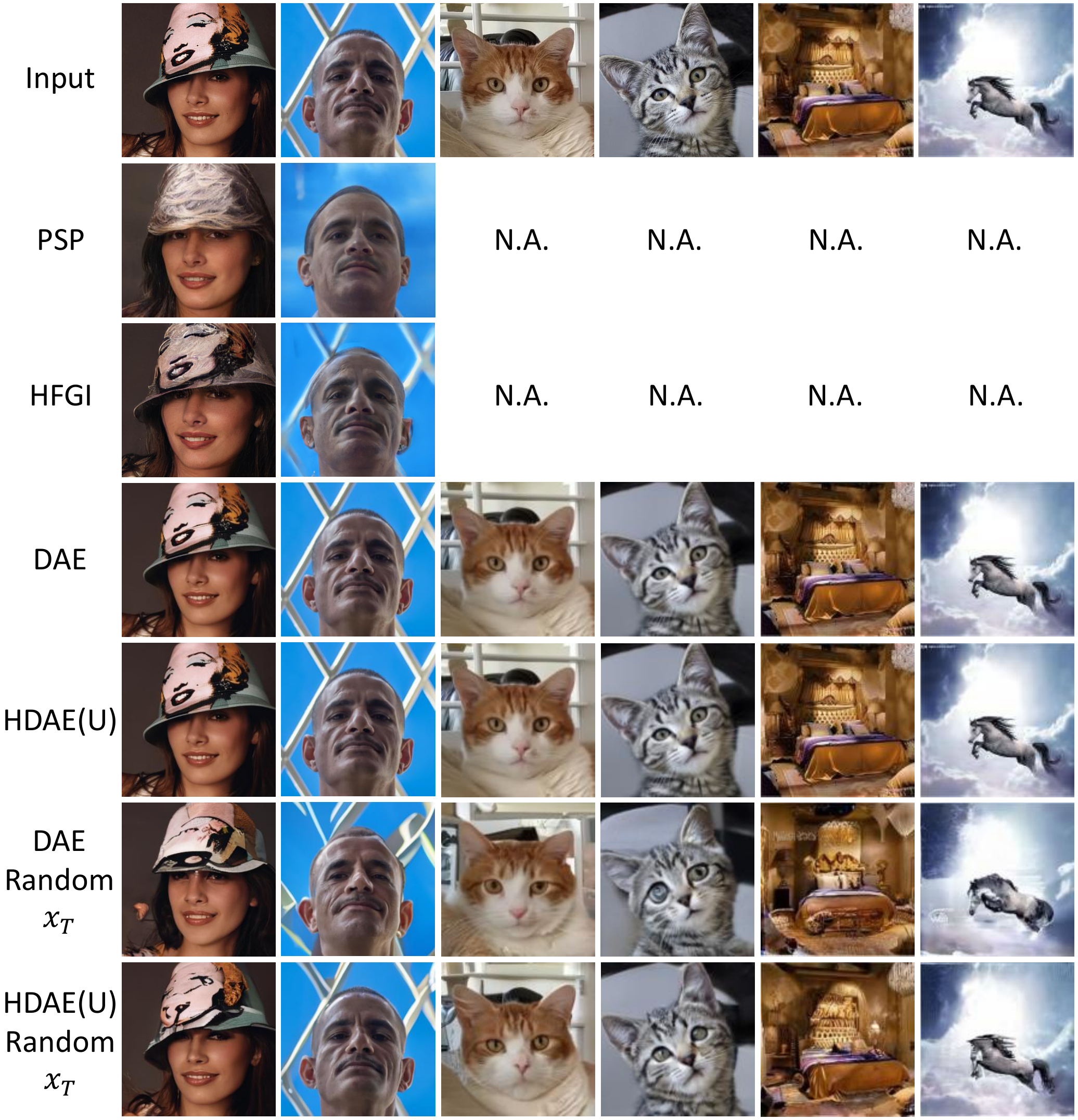}
    \caption{
        \textbf{Quantitative results of image reconstruction.}
        We evaluate the image reconstruction experiments on the face, cat, bedroom, and horse images.
        HDAE(U) with random stochastic code $x_T$ (7th row) preserves more details in backgrounds, appearance, expressions, and identity information than DAE with random stochastic code $x_T$ (6th row).
        HDAE(U) and DAE with encoded stochastic code $x_T$ (4th and 5th row) attain near-perfect reconstruction.
        “N.A.” denotes not applicable for this model.
    }
    \vspace{-0.4cm}
    \label{Reconstruct}
\end{figure}

\subsection{Image Reconstruction}

\noindent\textbf{Setup.}
Image reconstruction quality reflects how well the latent representations encode the image information, especially the details.
We evaluate the image reconstruction quality of different approaches: GAN inversion methods (StyleGAN2~\cite{Stylegan2} with \begin{math}\mathcal{W}\end{math} space, pretrained StyleGAN2 with \begin{math}\mathcal{W+}\end{math} space, VQ-GAN~\cite{VQGAN}, HFGI~\cite{HFGI}), VAE-based method (VQ-VAE2~\cite{VQVAE2}), and diffusion-based methods (DDIM~\cite{DDIM}, and DAE~\cite{DAE}, and our HDAE(U)). 
All these models are trained on FFHQ~\cite{Stylegan} and tested on 30,000 images from CelebA-HQ~\cite{ProGAN}.
To validate the potential of our model beyond the face domain, we show image reconstruction results on cat datasets by training the models on LSUN~\cite{Lsun} Cat and testing on the AFHQ Cat~\cite{Starganv2}.
In addition, we show our image reconstruction results on LSUN Bedroom and LSUN Horse datasets.
DDIM, DAE, and HDAE(U) are trained on images of size $128 \times 128$ and we use $T=100$ for inference. 
Structural Similarity SSIM, Pixel-wise MSE, and perceptual quality metric LPIPS~\cite{LPIPS} are adopted to evaluate the image reconstruction quality.
To illustrate what information is encoded by the stochastic encoder, we show the images reconstructed from DAE and HDAE(U) with their corresponding $x_T$ as well as random $x_T$ for comparison.

\noindent\textbf{Quantitative results.}
The results Tab.~\ref{imgrectab} demonstrate that: (1) HDAE(U) achieves near-perfect image reconstruction performance, outperforming previous GAN inversion methods, VAE-based approaches, and diffusion-based approaches.
(2) Comparing image reconstruction results with the $x_T$ encoded by a stochastic encoder and random $x_T$, some detail-related information is encoded by the stochastic encoder.
(3) The image reconstruction performance degrades more for DAE than HDAE(U) when replacing $x_T$ encoded by the stochastic encoder with a random $x_T$, indicating that in HDAE(U) more information is encoded in the semantic encoder and less information is encoded by the stochastic encoder.


\begin{figure}[!t]
    \centering
    \includegraphics[width=0.97\linewidth]{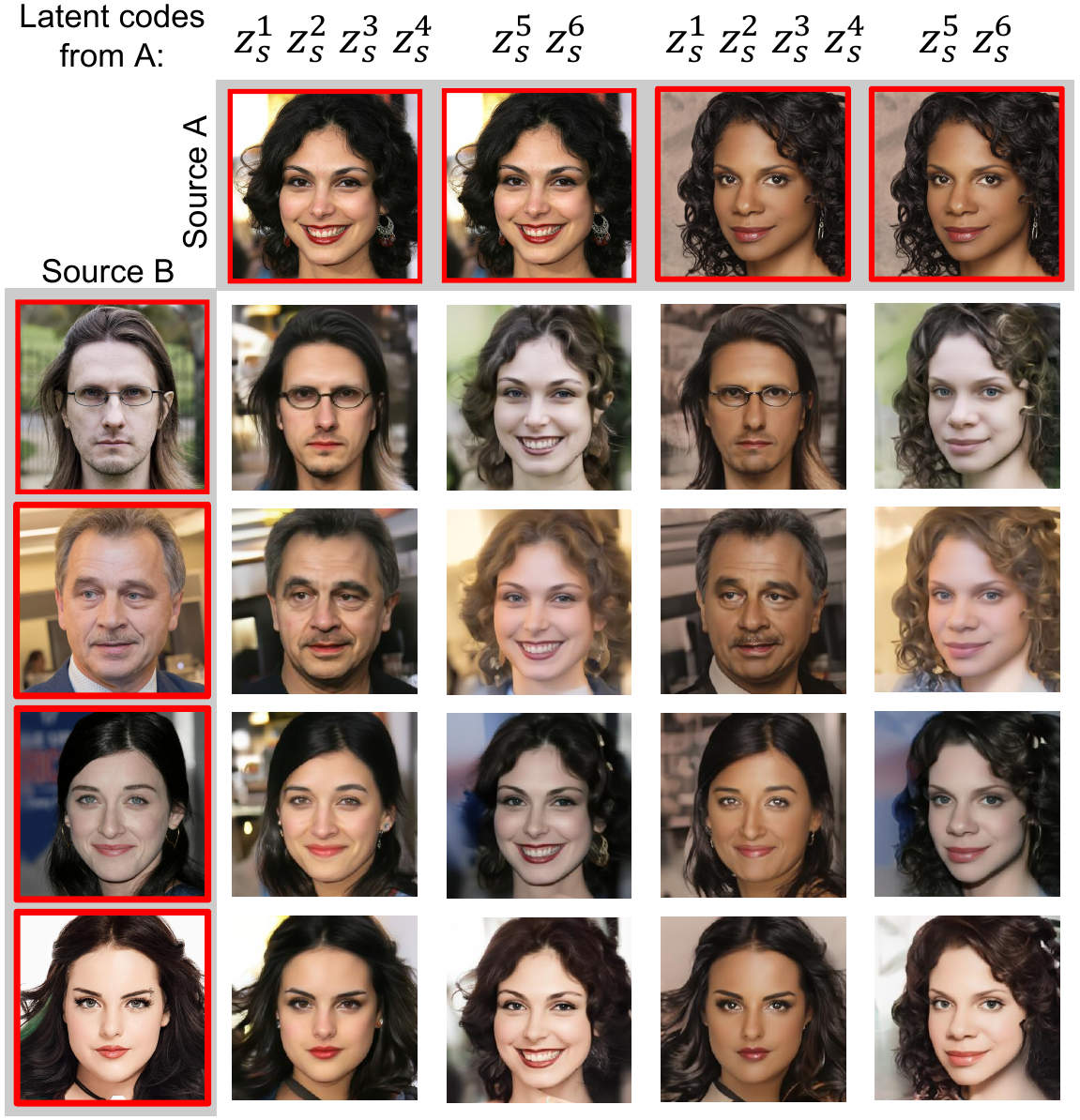}
    \caption{
        \textbf{Style mixing results with hierarchical latent space.}
        $z_s^1$,$z_s^2$,$z_s^3$,$z_s^4$ represent the low-level latent codes and $z_s^5$,$z_s^6$ represent the high-level latent codes.
        Given the real images in red boxes, we can mix the high-level latent codes from source A(B) and low-level latent codes from source B(A).
    }
    \vspace{-0.4cm}
    \label{interpolation}
\end{figure}
\begin{figure*}[!t]
    \centering
    \includegraphics[width=0.88\linewidth]{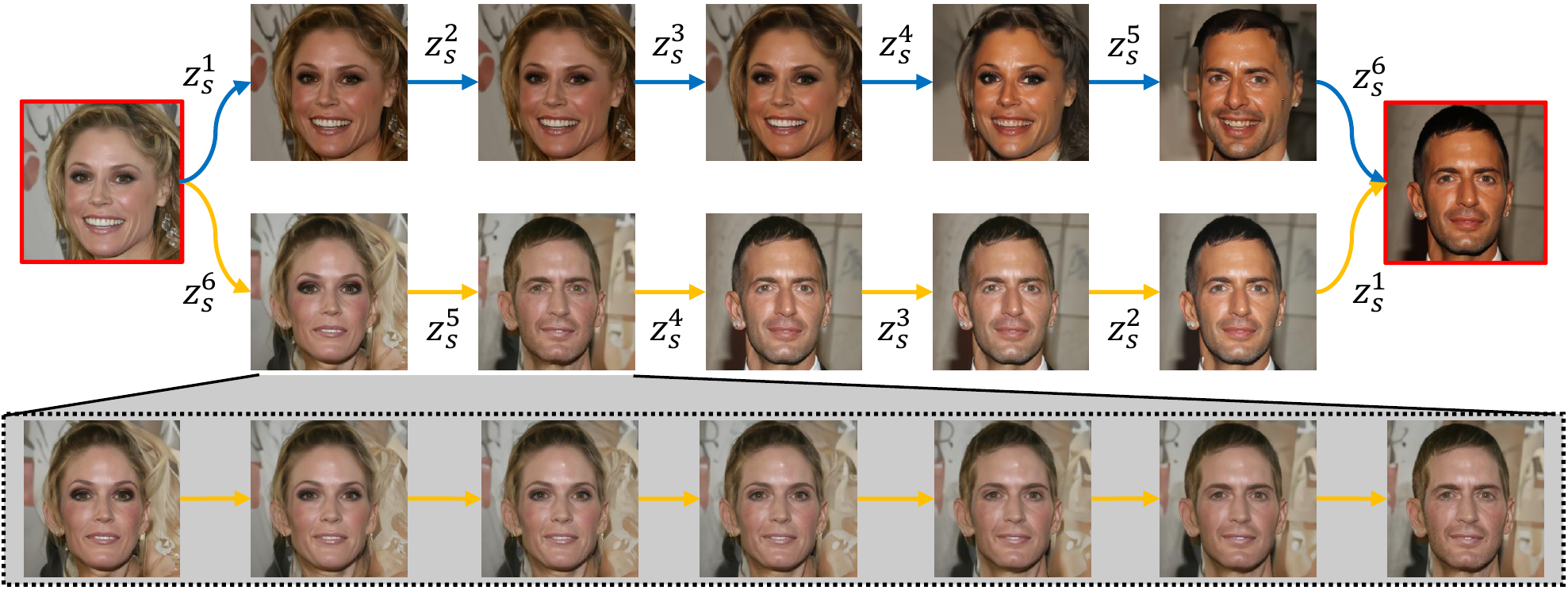}
    \caption{
        \textbf{Controllable image latent space interpolation along different paths.}
        Given the real images in red boxes,
        the first path shows the image interpolation changing from low-level latent codes to high-level latent codes. The second path shows the reverse interpolation process from high-level to low-level. The third path shows the smooth interpolation between two images by smoothly interpolating $z_s^5$.
    }
    \vspace{-0.25cm}
    \label{pro_interpolation}
\end{figure*}

\noindent\textbf{Qualitative results.}
The qualitative results of image reconstruction are shown in Fig.~\ref{Reconstruct}. Previous GAN inversion approaches PSP~\cite{PSP} and HFGI~\cite{HFGI} cannot preserve the background and details of the original images, while DAE and HDAE(U) with encoded $x_T$ reconstruct images nearly identical to the input images.
Comparing \textit{DAE with random $x_T$} with \textit{HDAE(U) with random $x_T$}, we find that HDAE(U) with random $x_T$ preserves the background and details better. 
DAE with random $x_T$ fails to reconstruct the details, such as the eyes of the cat, the interior decoration of the bedroom, and the legs of the horse.
This observation indicates that the semantic encoder of HDAE(U) encodes more comprehensive features, allowing detail-preserving applications such as image manipulation.
More examples are in the appendix.

\noindent\textbf{Human perceptual evaluation.}
We conduct a human perceptual evaluation, where users are asked to vote for the image reconstruction quality of HDAE(U) with random $x_T$ and DAE with random $x_T$ in Tab.~\ref{userstudytab}.
We collect 450 votes from 15 participants, and the results are shown in Tab.~\ref{userstudytab}.
Users clearly prefer the image reconstruction results by our HDAE(U) over DAE.
More details are in the appendix.


\subsection{Interpreting the Hierarchical Latent Space}
The plot in Fig.~\ref{clsheatmap} indicates a strong correlation between feature levels and attributes.
For instance, ``brown hair'' is correlated with the low-level feature layers 1 and 2, while ``smiling'' is correlated with the high-level feature layer 4.
It reveals the hierarchical latent space where different layers represent different abstraction levels of the representations. 
We visualize the latent space hierarchy by style mixing and controllable image interpolation experiments.

\noindent\textbf{Style mixing.}
To interpret and visualize the hierarchical latent space, we mix the  high-level latent codes $z_s^5,z_s^6$ from source image A(B) and the low-level latent codes $z_s^1, z_s^2, z_s^3, z_s^4$ from image B(A), and use the mixed latent codes to generate a new image, as shown in Fig.~\ref{overview}.
Fig.~\ref{interpolation} shows a clear hierarchy of the latent space. $z_s^1, z_s^2, z_s^3, z_s^4$ control the spatial details such as background, color, and lighting, and $z_s^5,z_s^6$ control the high-level semantics and image structure such as pose, gender, face shape, and eyeglasses.

\noindent\textbf{Controllable image interpolation.}
Image interpolation based on semantic codes is a common way to visualize and verify the properties of the latent space.
With our hierarchical latent space, we can control different paths of image interpolation, as shown in Fig.~\ref{pro_interpolation}.
Given the leftmost and rightmost real images in red boxes,
in the first row, we interpolate from left to right by changing low-level features first and then high-level features.
In the second row, the interpolation is conducted in a reverse way, from high-level features to low-level features.
In the third row, we illustrate the continuous changes between the two images in the second row.
Results indicate that 
$z_s^1,z_s^2,z_s^3$ control lighting and color, $z_s^4$ controls background, $z_s^5$ controls gender and $z_s^6$ controls pose.


\begin{figure}[!t]
    \centering
    \includegraphics[width=0.98\linewidth]{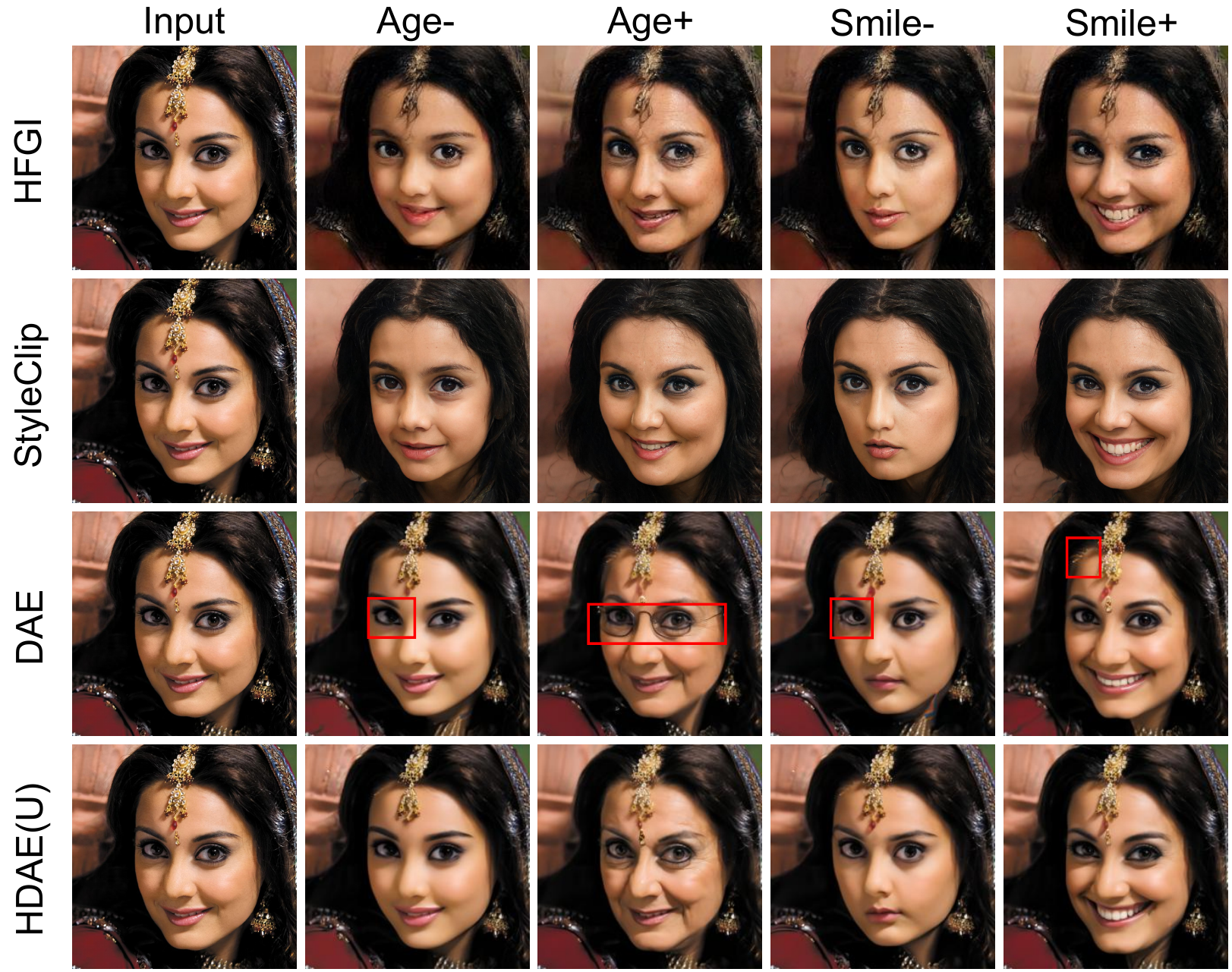}
    \caption{
        \textbf{Comparisons on real image manipulation between HFGI~\cite{HFGI}, StyleClip~\cite{Styleclip}, DAE~\cite{DAE} and HDAE(U).}
        HDAE(U) preserves the details (\eg, face identity, background, and the forehead pendant) in the original image better than other approaches.
    }
    \vspace{-0.6cm}
    \label{facemanipulate}
\end{figure}

\subsection{Image Manipulation}\label{subsec:disentangled}

\begin{figure*}[!h]
    \centering
    \includegraphics[width=0.96\linewidth]{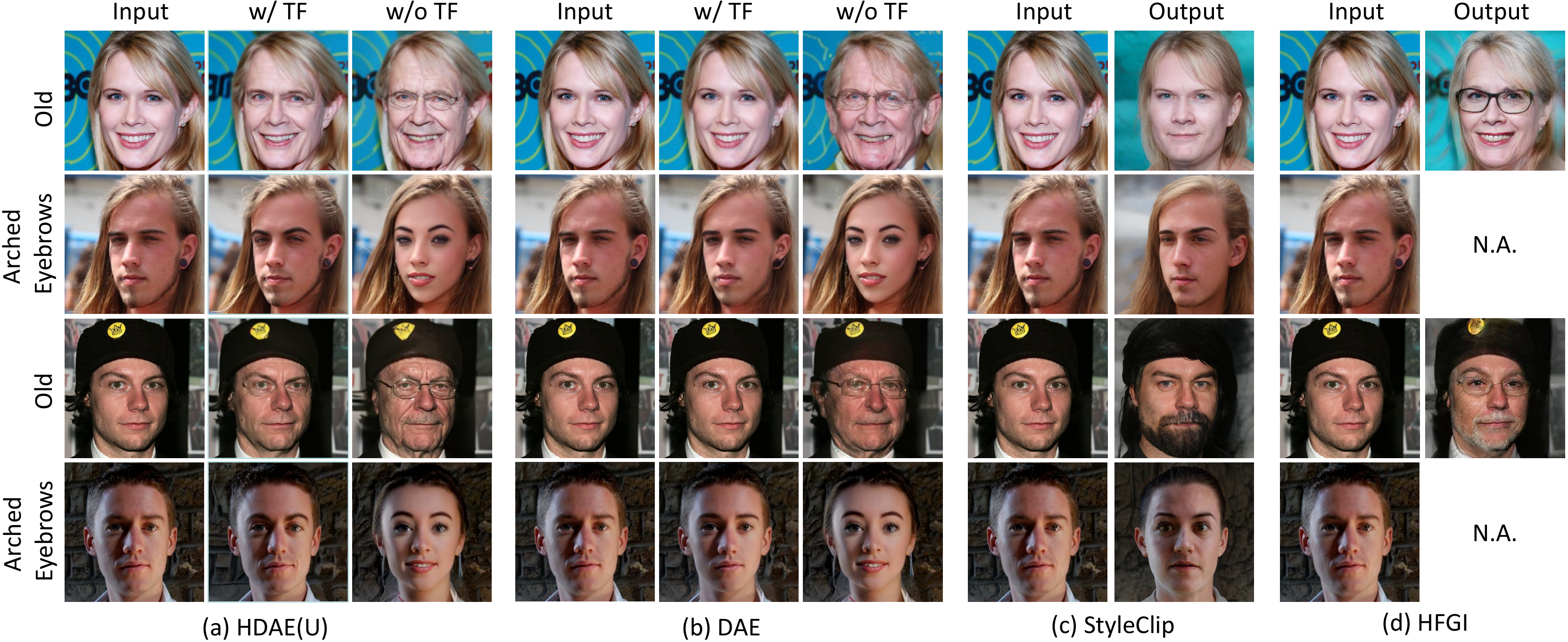}
    \caption{
        \textbf{Disentangled attribute manipulation results.}
        HDAE(U) is trained on FFHQ~\cite{Stylegan} with image resolutions $256 \times 256$.
        We preserve the top $k$ largest values for the truncated features.
        ``w/ TF" denotes using truncated features and $k=24$ gives the best manipulation results in terms of attribute disentanglement (see appendix for ablation on $k$).
        ``w/o TF" 
        is the comparison experiment without truncated features, where we can observe severe attribute entanglements: old - wearing eyeglasses, arched eyebrows - female, female - makeup.
        “N.A.” denotes not applicable for this model.
    }
    \vspace{-0.25cm}
    \label{topkmanipulate}
\end{figure*}

\noindent\textbf{Detail-preserving image manipulation with HDAE.}
With the linear attribute classifiers for the latent codes, we can edit real images by manipulating the semantic latent codes with the classifier direction.
Fig.~\ref{facemanipulate} shows the image manipulation results on HFGI~\cite{HFGI}, StyleClip~\cite{Styleclip}, DAE~\cite{DAE}, and our HDAE(U).
DAE and HDAE(U) are trained on FFHQ, and the linear attribute classifiers are trained on the CelebA-HQ.
As demonstrated in Fig.~\ref{facemanipulate},
our HDAE(U) preserves the details (\eg, background, face identity, and the forehead pendant) of the input image better than other approaches.
The image manipulation results demonstrate that the representations learned by our HDAE are rich and semantically meaningful. 
More examples can be found in the appendix.

\noindent\textbf{Disentangled image manipulation with truncated features.}
As introduced in Sec.~\ref{subsec:disentangled}, we leverage the truncated features for disentangled image manipulation.
We compare the qualitative results of image manipulation with truncated features ($k=24$) and without truncated features ($k=3072$) on DAE and HDAE(U).
We also compare with GAN-based methods HFGI~\cite{HFGI} and StyleClip~\cite{Styleclip}.
As shown in Fig.~\ref{topkmanipulate}, for the na\"ive manipulation without truncation, ``old'' is entangled with ``eyeglasses'', ``arched eyebrows'' is entangled with ``female'', and ``female'' is entangled with ``makeup''.
Our HDAE(U) with truncated features successfully disentangles those attributes and gives the best results compared with other methods.
The truncated features with $k=24$ effectively disentangle the attributes for manipulation, and the attributes become more entangled as $k$ increases (from $k=24$ to $k=3072$).
An ablation study of $k$ and more examples are in the appendix.


\noindent\textbf{Human perceptual evaluation.}
We conduct user perceptual evaluations on the image manipulation and disentangled image manipulation results of HDAE(U), DAE, HDAE(U) with truncated features and DAE with truncated features.
For image manipulation, we collect 1,575 votes from 15 participants.
For disentangled image manipulation, we collect 2,100 votes from 15 participants.
Results in Tab.~\ref{userstudytab} indicate that HDAE(U) performs better than DAE, and that HDAE(U) with truncated features performs better than HDAE(U), DAE, and DAE with truncated features.
Details are in the appendix.

\noindent\textbf{Discussion.}
The hierarchical latent space and truncation-based approach are orthogonal approaches to improve image manipulation from different perspectives.
The feature hierarchy improves image manipulation by providing a comprehensive and semantically meaningful latent space. The richness of the latent space ensures detail preservation for image manipulation.
On the other hand, the truncation-based approach empowers disentangled image manipulation.
Therefore, HDAE(U) with truncated features shows the best detail-preserving and disentangled image manipulation results.

\subsection{Other Applications}\label{subsec:other_applications}
\begin{figure}[!t]
    \centering
    \includegraphics[width=0.98\linewidth]{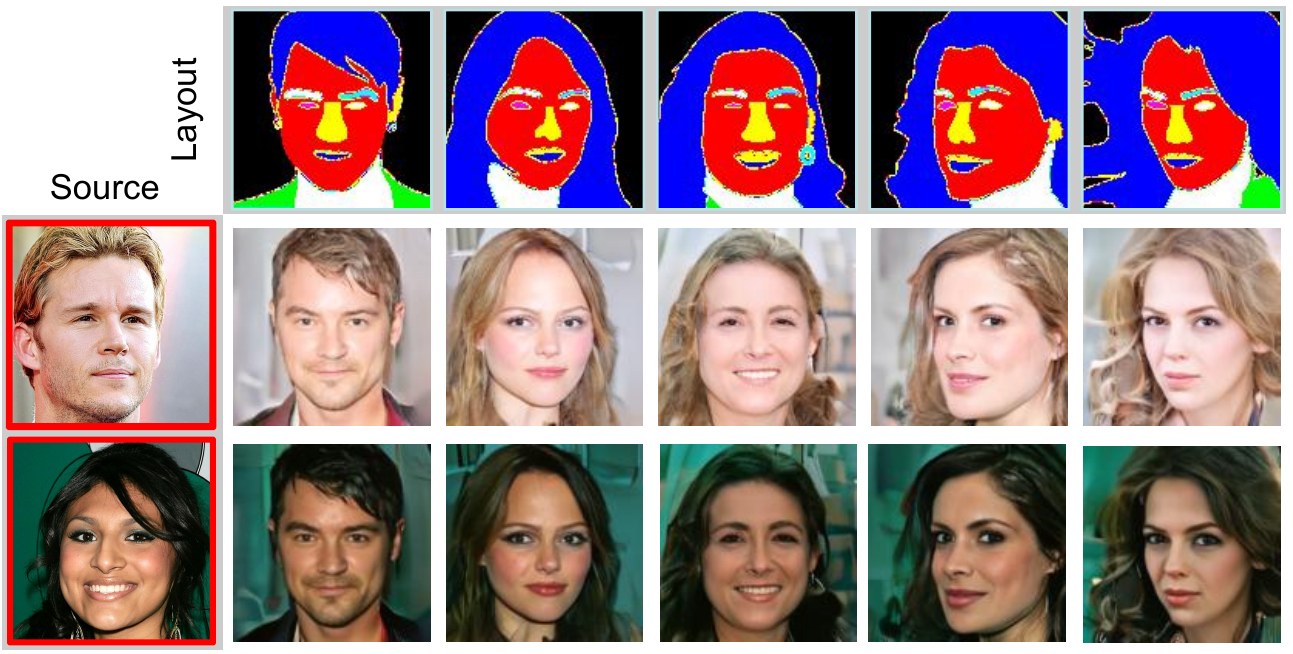}
    \caption{
        \textbf{Multi-modal semantic image synthesis.} Our HDAE conducts style mixing with the layout information from the semantic layout map and details from the source image.
    }
    \vspace{-0.35cm}
    \label{s2i}
\end{figure}

\noindent\textbf{Multi-modal semantic image synthesis.}
We train an extra layout encoder that maps the semantic label map into the latent space of our HDAE.
Our HDAE can synthesize images based on the high-level features from a semantic label map and the low-level features from a real image. In this way, we can control the layout of the generated images with the label map, and control the image style and details with the image, as shown in Fig.~\ref{s2i}.

\noindent\textbf{Unconditional image synthesis.}
By training a latent DDIM model to predict the latent codes, our model can be leveraged for unconditional image synthesis. Results and more details can be found in the appendix.

\label{sec:experiments}

\section{Conclusion}

We present Hierarchical Diffusion Autoencoders (HDAE) which leverages the feature hierarchy to build a hierarchical latent space for diffusion models. 
The latent representations are rich and comprehensive, with a coarse-to-fine hierarchy.
We further propose a novel disentangled attribute manipulation approach with truncated features.
Extensive experiments and applications on image reconstruction, style mixing, controlled image interpolation, disentangled image editing, and multimodal semantic image synthesis are conducted to validate the effectiveness of our approach.
\label{sec:conclusion}

\newpage

{\small
\bibliographystyle{ieee_fullname}
\bibliography{section/11_references}
}

\clearpage 
\appendix
\section*{\label{sec:appendix} Appendix}

In Section~\ref{sec:Architectures}, we provide details on the architecture design and training setups. In Section~\ref{sec:More Experiments}, we provide more experiments and visual results to demonstrate the effectiveness of Hierarchical Diffusion Autoencoders.

\section{Architectures}
\label{sec:Architectures}
\begin{figure}[h]
    \centering
    \includegraphics[width=0.94\linewidth]{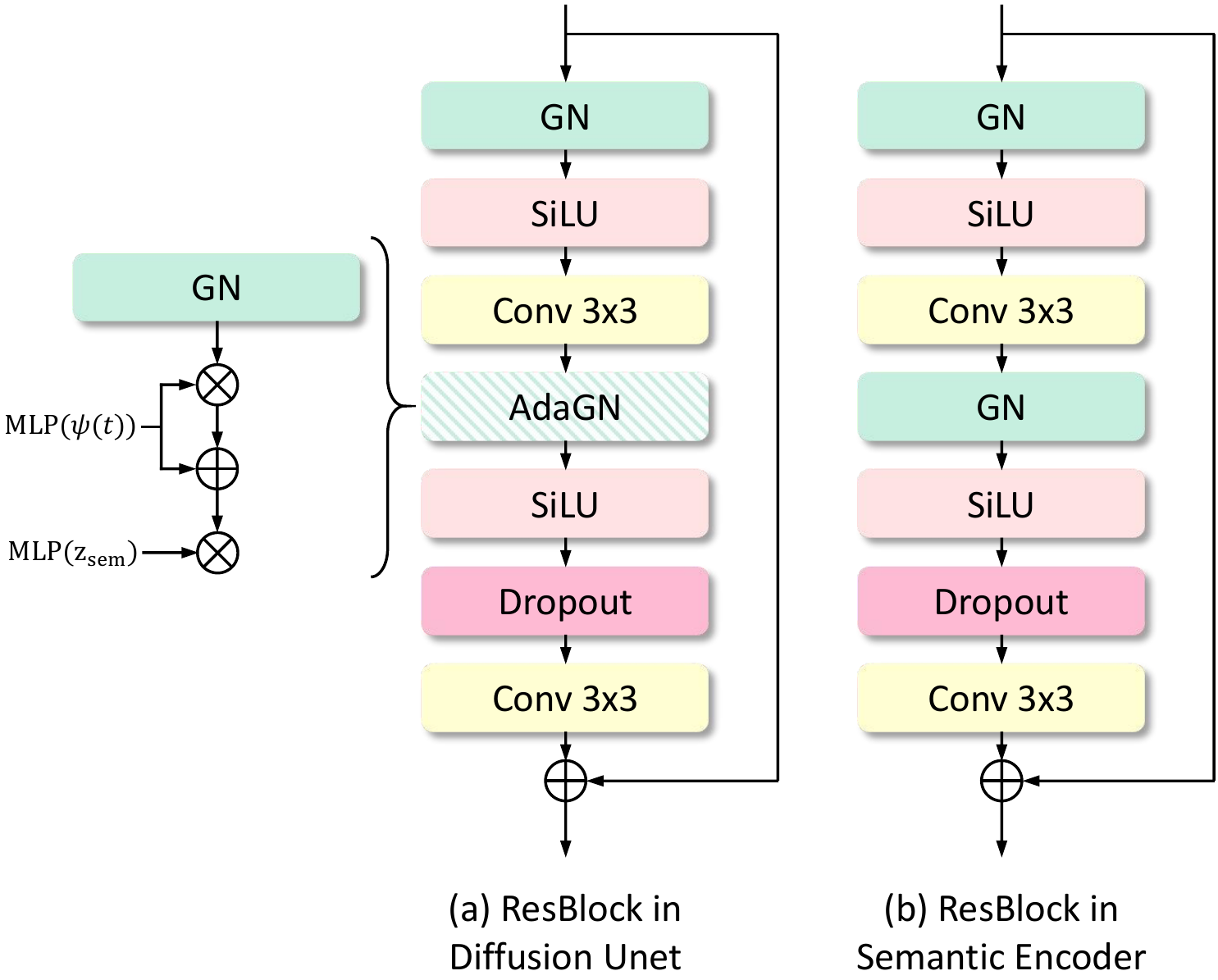}

    \caption{
        \textbf{Architecture overview of Hierarchical Diffusion Autoencoders.}
    }

    \label{arch_sup}
\end{figure}

Our Hierarchical Diffusion Autoencoders adopts the same diffusion-based decoder architecture with Diffusion Autoencoders~\cite{DAE} \footnote{\url{https://github.com/phizaz/diffae}}.
The network architectures of Resblocks in the diffusion U-net and semantic encoder are shown in Fig.~\ref{arch_sup}.
Following Diffusion Autoencoders, the timestep embeddings and semantic codes are fed into the diffusion-based decoder as conditions with AdaGN layers.
For the fairness of the experiments, we try to keep most of the hyperparameters consistent with Diffusion Autoencoders~\cite{DAE}.

The network architecture, hyperparameters, and training parameters of Hierarchical Diffusion Autoencoders are shown in Tab.~\ref{tab_arch_sup}.


\section{Experiments}
\label{sec:More Experiments}
\subsection{Evaluating the Fidelity of Image Manipulation}
An important evaluation criterion for image manipulation is the fidelity of image manipulation, \ie, how well the manipulated images preserve the details of the original images.

\begin{figure}[t]
    \centering
    \includegraphics[width=0.94\linewidth]{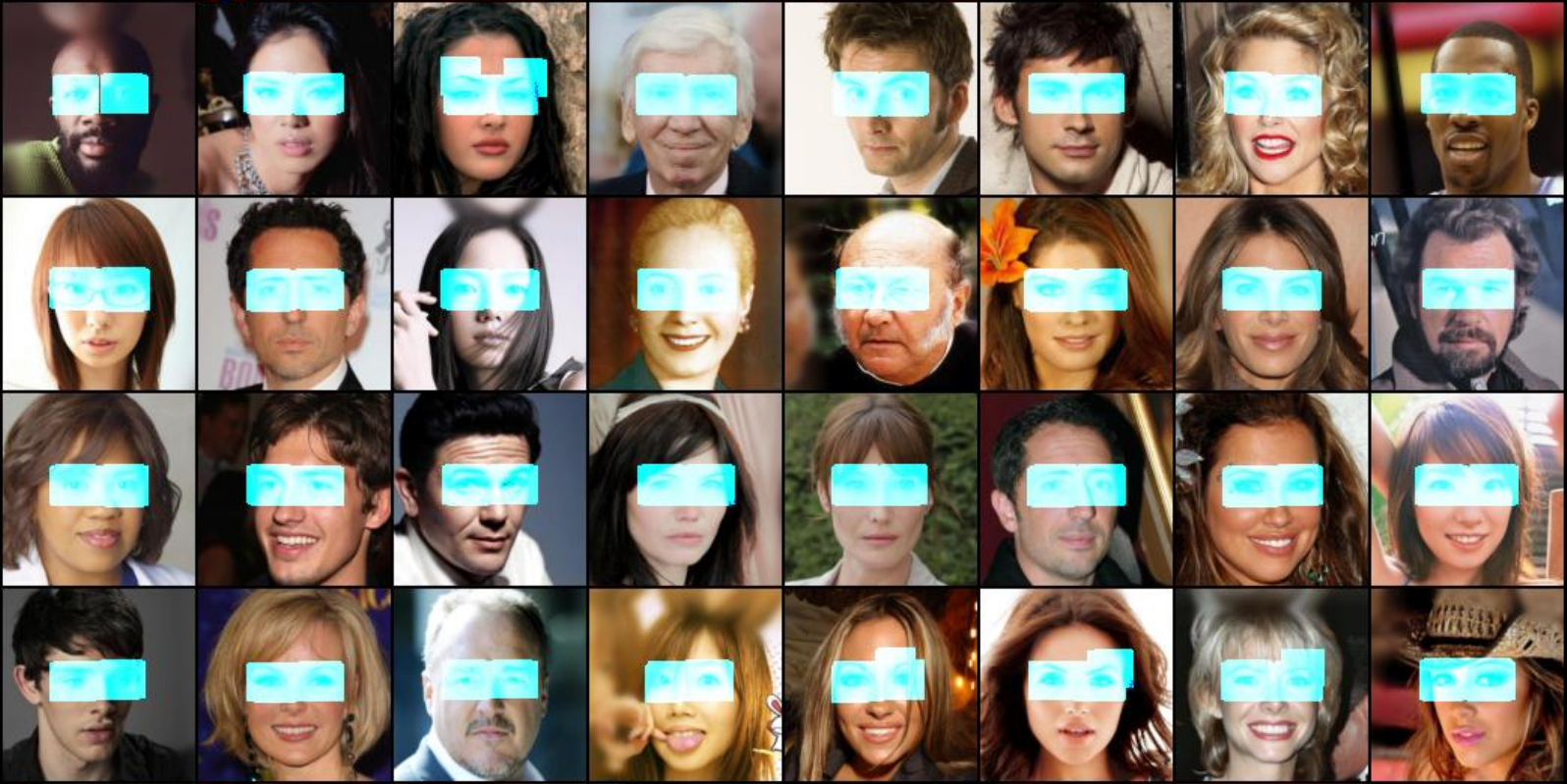}
    \caption{
        \textbf{The visualization of the dilated binary masks generated from the segmentation maps of the eyes.}
        We highlight the area to be edited, and the remaining non-highlighted areas should be consistent before and after editing in our image manipulation with high fidelity task.
    }
    \label{fim_sup}
\end{figure}

We conduct the experiments on the face datasets CelebA-HQ~\cite{ProGAN}.
Given an image $I_0$ and an attribute $a$ to manipulate, 
we first extract a binary mask $I_{bin}^a$ to indicate the rough regions related to the attribute, by dilating the segmentation map of the attribute.
$I_{bin}^a$ highlights the area to be manipulated ($I_{bin}^a = 1$ in this area), and $1-I_{bin}^a$ highlights the area that should be consistent before and after editing, as shown in Fig.~\ref{fim_sup}.
Then we apply our Hierarchical Diffusion Autoencoders to manipulate the original image $I_0$ with the specific attribute $a$ to get the manipulated image $I_m^a$.
Finally, we can compute the LPIPS~\cite{LPIPS} and MSE metrics between $I_0*(1-I_{bin}^a)$ and $I_m^a*(1-I_{bin}^a)$.
We calculated the average value of MSE and LPIPS from $\alpha=-0.5$ to $\alpha=0.5$ ($\alpha$ is the coefficient of editing direction, which controls the degree of editing) and Tab.~\ref{tab_fim_sup} reports the results of DAE, HDAE(U) and HDAE(U) with disentangled attribute manipulation. The results demonstrate that our HDAE(U) preserves the details better than HDAE, and the disentangled manipulation with truncated features achieves the best performance in terms of fidelity.

\begin{table*}[t]

\resizebox{1\linewidth}{!}{
    \begin{tabular}{l|ccccc}
        \toprule
        \textbf{Hyperparameter}  & \textbf{DAE (FFHQ 128)} & \textbf{DAE(2560) (FFHQ 128)} & \textbf{DAE(U) (FFHQ 128)} & \textbf{HDAE(E) (FFHQ 128)} & \textbf{HDAE(U) (FFHQ 128)}\\
        \midrule
        Batch size                   &  128               &128                   & 128               & 128               & 128              \\
        Base channels                & 128                & 128                  & 128               & 128               & 128               \\
        Channel multipliers          & {[}1,1,2,3,4{]}    &  {[}1,1,2,3,4{]}     &  {[}1,1,2,3,4{]}  & {[}1,1,2,3,4{]}   & {[}1,1,2,3,4{]}   \\
        Images trained               & 130M               & 130M                 & 130M              & 130M              & 130M             \\
        Encoder base ch              & 128                & 128                  & 128               & 128               & 128              \\
        Encoder ch. mult.            & {[}1,1,2,3,4,4{]}  & {[}1,1,2,3,4,4{]}    & {[}1,1,2,3,4,4{]} & {[}1,1,2,3,4,4{]} & {[}1,1,2,3,4,4{]} \\
        Decoder base ch              & -                  & -                    & 128               &        -          & 128               \\
        Decoder ch. mult.            &-                   &-                     &{[}1,1,2,3,4,4{]}  & -                 & {[}1,1,2,3,4,4{]}  \\
        Attention resolution         & {[}16{]}           & {[}16{]}             & {[}16{]}          & {[}16{]}          & {[}16{]}          \\
        Latent code dim              & 512                & 2560                 & 512               & 2560              & 2560             \\
        $\beta$ scheduler            & Linear             & Linear               & Linear            & Linear            & Linear            \\
        Learning rate                & \multicolumn{5}{c}{1e-4}                                                         \\
        Optimizer                    & \multicolumn{5}{c}{Adam (no weight decay)}                                       \\
        Training $T$                 & \multicolumn{5}{c}{1000}                                                         \\
        Diffusion loss               & \multicolumn{5}{c}{MSE with noise prediction $\vec{\epsilon}$}                                    \\
        Diffusion var.               & \multicolumn{5}{c}{Not important for DDIM} \\       
        \midrule
        Parameters & 122.59M & 190.6M & 160.94M & 154.62M & 189.15M \\
        \bottomrule
    \end{tabular}
}
\caption{\textbf{Network architecture and training hyperparameters of hierarchical diffusion autoencoder.}}

\label{tab_arch_sup}
\end{table*}

\begin{table*}[t]
\renewcommand\arraystretch{1.4}
\resizebox{1\linewidth}{!}{
\begin{tabular}{l|cccccccc}
    \toprule[1.2pt]
\textbf{Metric}              & \multicolumn{8}{c}{\textbf{LPIPS}($\downarrow$)}                                                                                      \\ \hline
\textbf{Attribute}           & \textbf{eyeglasses} & \textbf{mouth\_slightly\_open} & \textbf{big\_lips} & \textbf{big\_nose} & \textbf{mustache} & \textbf{bags\_under\_eyes} & \textbf{arched\_eyebrows} & \textbf{gray\_hair} \\ \hline
DAE~\cite{DAE}                 & 0.23267  & 0.18339             & 0.10974  & 0.15190   & 0.18660   & 0.20546         & 0.18365         & 0.31655  \\
HDAE(U)             & 0.22098  & 0.16874             & 0.10437  & 0.13695  & 0.18223  & 0.19380          & 0.17504         & 0.28918  \\
HDAE(U) Disentangled &   \textbf{0.18237}       &    \textbf{0.13382}                 &  \textbf{0.07329}        &    \textbf{0.10872}      &    \textbf{0.14892}      &       \textbf{0.12983}          &      \textbf{0.13568}           &   \textbf{0.22260}       \\ \midrule[1.2pt]  \midrule[1.2pt]
\textbf{Metric}              & \multicolumn{8}{c}{\textbf{MSE}($\downarrow$)}                                                                                        \\ \hline
\textbf{Attribute}           & \textbf{eyeglasses} & \textbf{mouth\_slightly\_open} & \textbf{big\_lips} & \textbf{big\_nose} & \textbf{mustache} & \textbf{bags\_under\_eyes} & \textbf{arched\_eyebrows} & \textbf{gray\_hair} \\ \hline
DAE~\cite{DAE}                 & 0.01105  & 0.00906             & 0.00341  & 0.00388  & 0.01060   & 0.00952         & 0.00588         & 0.08564  \\
HDAE(U)             & 0.01011  & 0.00804             & 0.00262  & 0.00304  & 0.00823  & 0.00843         & 0.00548         & 0.07369  \\
HDAE(U) Disentangled &   \textbf{0.00771}       &         \textbf{0.00593}            &     \textbf{0.00126}     &     \textbf{0.00165}     &  \textbf{0.00517}        &         \textbf{0.00595}        &    \textbf{0.00389}            &    \textbf{0.04739}      \\     \bottomrule[1.2pt]
\end{tabular}
}
\caption{\textbf{Evaluation of image manipulation fidelity.} ``HDAE(U) Disentangled'' means disentangled image manipulation with truncated feature using HDAE(U).}
\label{tab_fim_sup}
\end{table*}

\subsection{Image Reconstruction}

\begin{figure*}[!h]
    \centering
    \includegraphics[width=0.72\linewidth]{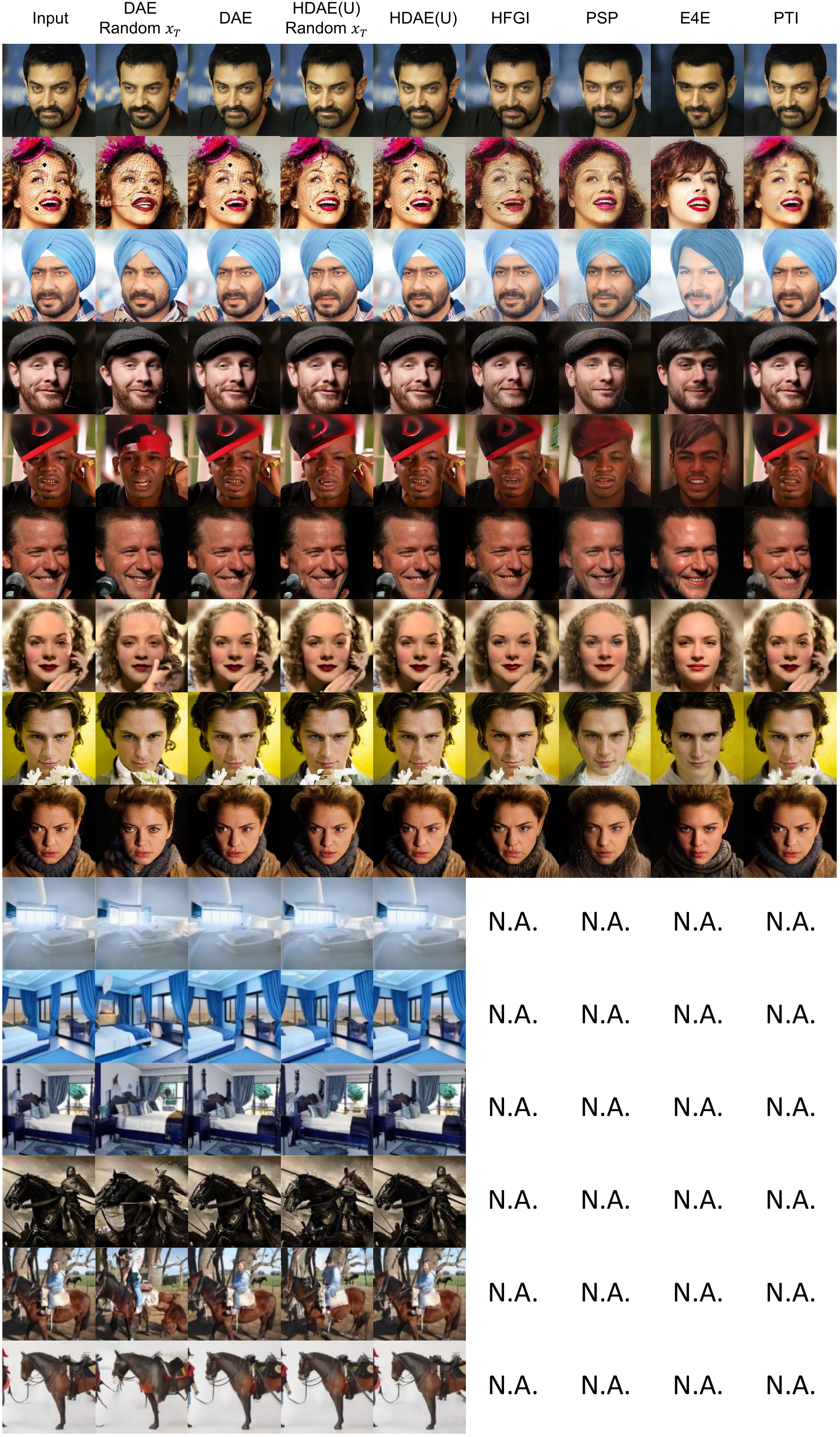}
    \caption{
        \textbf{Quantitative results of face and cat image reconstruction between HFGI~\cite{HFGI}, PSP~\cite{PSP}, E4E~\cite{Toc2021e4e}, PTI~\cite{roich2021pivotal}, DAE~\cite{DAE} and HDAE(U).}
    }
    \label{recon_sup}
\end{figure*}

We show the images reconstructed from DAE~\cite{DAE} and HDAE(U) with their corresponding $x_T$ as well as random $x_T$ for comparison.
We also show the results from some state-of-the-art GAN-based methods, such as HFGI~\cite{HFGI} and PSP~\cite{PSP}.
As shown in Fig. \ref{recon_sup}, our HDAE can achieve better reconstruction results than DAE with random $x_T$, indicating that our HDAE encodes richer and more comprehensive representations in the hierarchical semantic latent code.

\subsection{Image Manipulation}

\begin{figure*}[!h]
    \centering
    \includegraphics[width=0.75\linewidth]{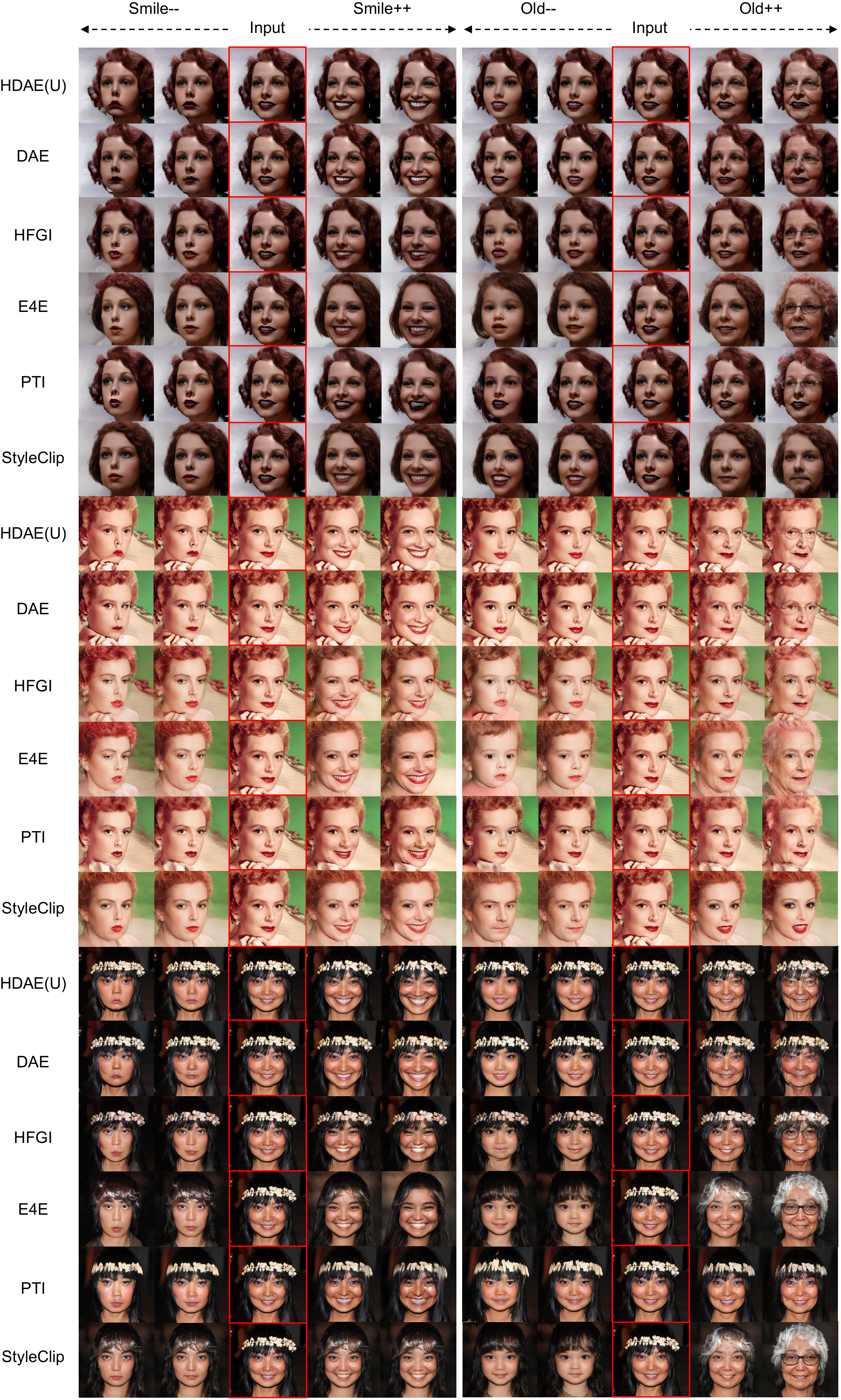}
    \caption{
        \textbf{Comparisons on real image manipulation between HFGI~\cite{HFGI}, E4E~\cite{Toc2021e4e}, PTI~\cite{roich2021pivotal}, StyleCLIP~\cite{Styleclip}, DAE~\cite{DAE} and HDAE(U).}
    }
    \vspace{0.8cm}
    \label{manipulate1_sup}
\end{figure*}

\begin{figure*}[!h]
    \centering
    \includegraphics[width=1\linewidth]{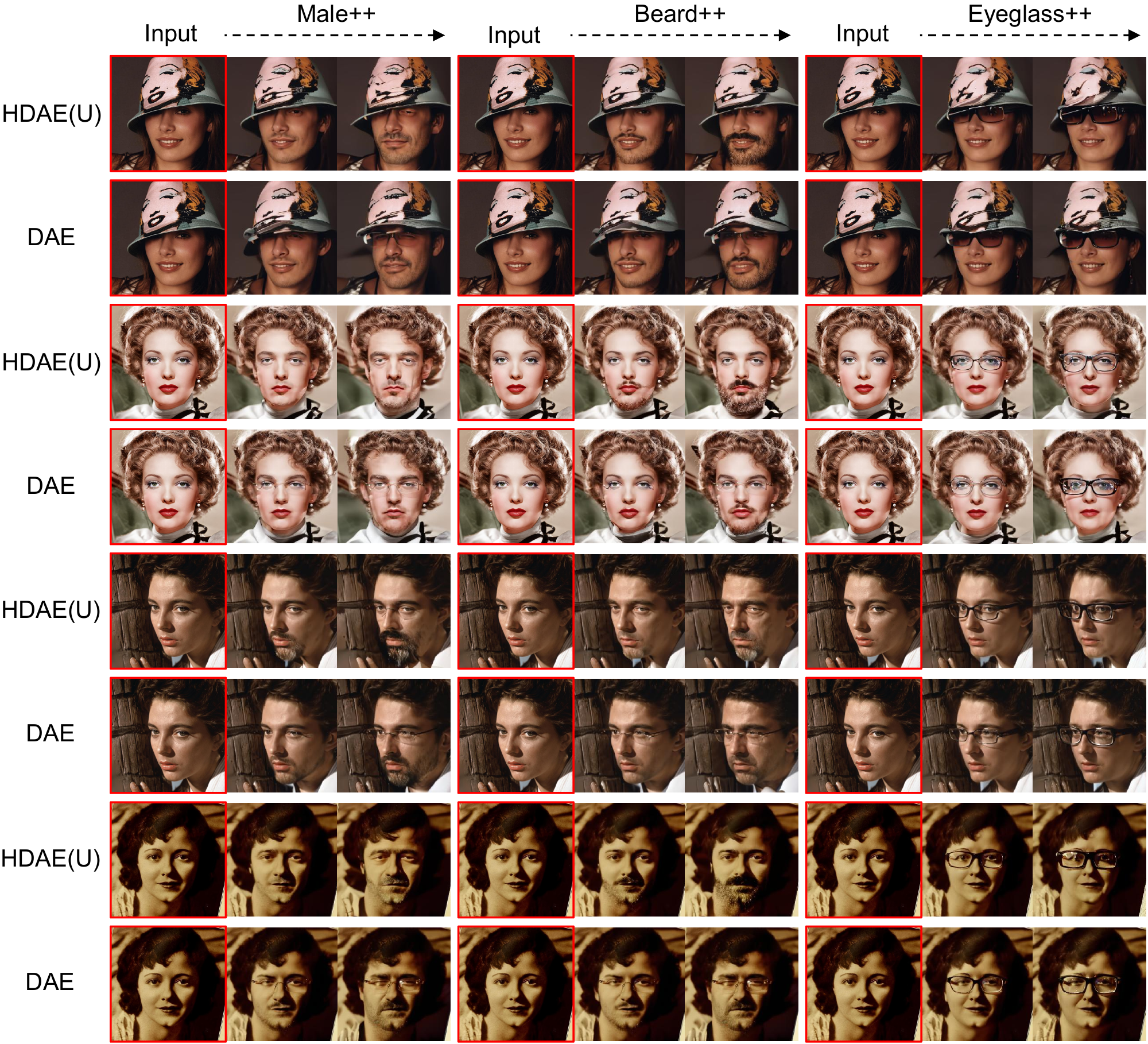}
    \caption{
        \textbf{Comparisons on real image manipulation between DAE~\cite{DAE} and HDAE(U).}
    }
    \vspace{5cm}
    \label{manipulate2_sup}
\end{figure*}

\noindent\textbf{Detail-preserving image manipulation.}
Fig.~\ref{manipulate1_sup} shows the visual results of image manipulation on real images with GAN inversion
approach HFGI~\cite{HFGI}, E4E~\cite{Toc2021e4e}, our baseline
DAE~\cite{DAE}, and our proposed model HDAE(U).
Fig.~\ref{manipulate2_sup} further compares the visual results of image manipulation on real images between DAE and HDAE(U).
The qualitative results demonstrate that our HDAE(U) is extremely good at preserving details for image editing, compared with previous approaches.

\begin{figure*}[!h]
    \centering
    \includegraphics[width=0.85\linewidth]{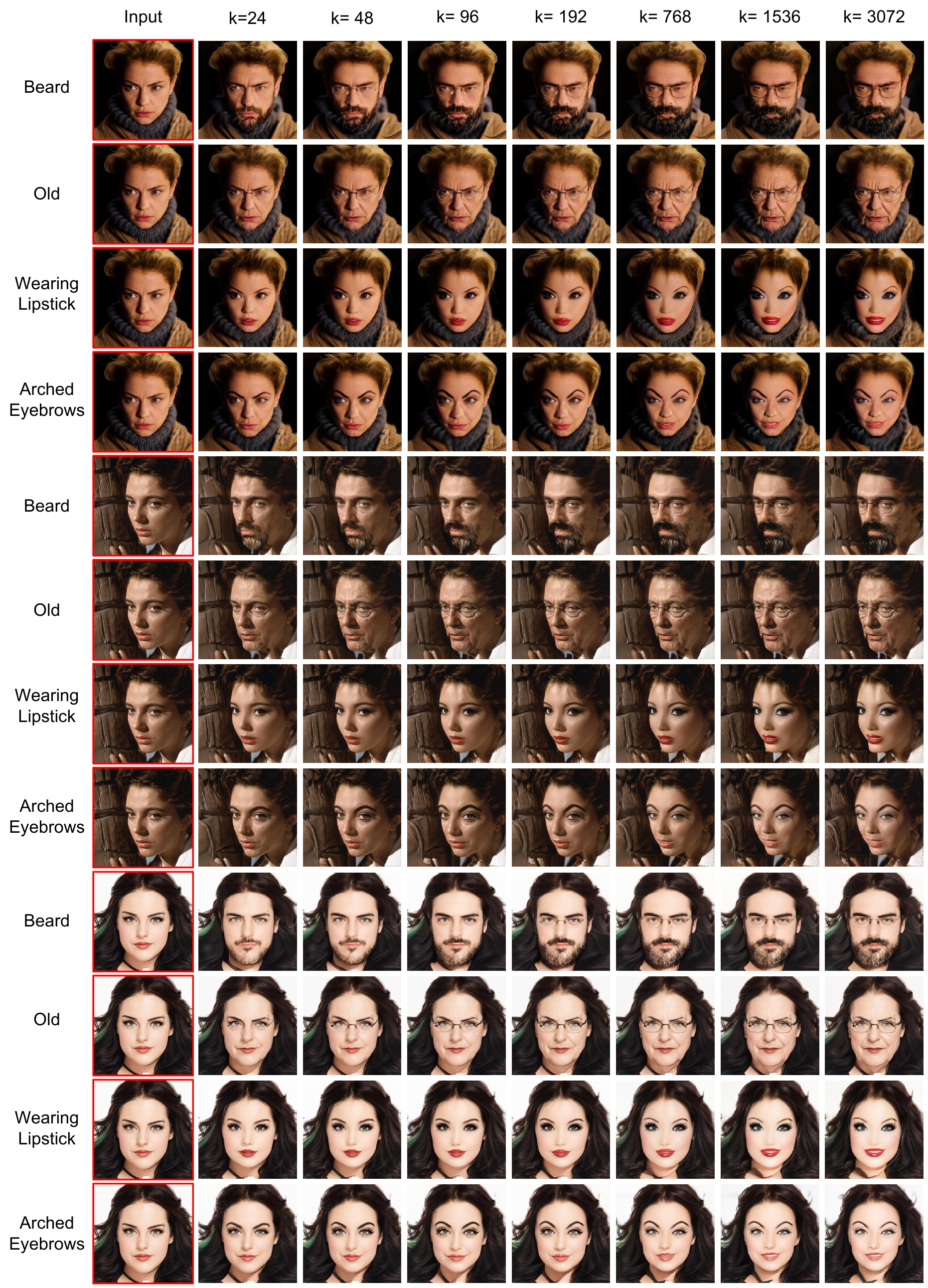}
    \caption{
        \textbf{Disentangled attribute manipulation results.}
    }
    \label{topk1_sup}
\end{figure*}

\begin{figure*}[!h]
    \vspace{-0.2cm}
    \centering
    \includegraphics[width=0.94\linewidth]{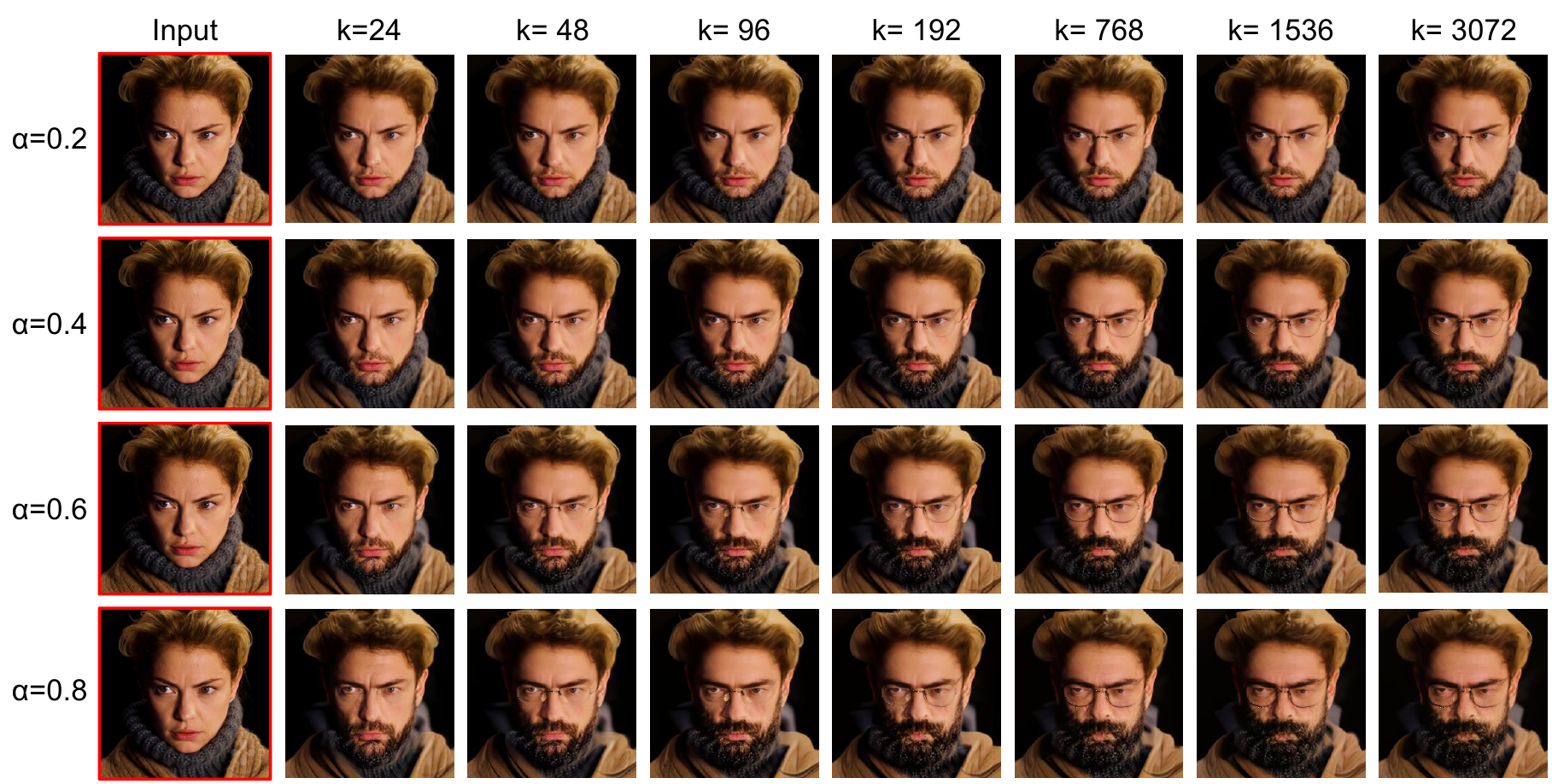}
    \caption{
        \textbf{Disentangled attribute manipulation results with different $\alpha$ and $k$.}
        As shown in the Figure, we manipulate the attribute of beard.
        And we can see that glasses will appear, as $k$ increases,
    }
    \label{topk2_sup}
\end{figure*}

\noindent\textbf{Disentangled image manipulation with truncated features.}
We show the qualitative results of image manipulation with different $\alpha$ (the weight of classifier direction) and $k$ ($k$ channels are preserved after truncation) in Fig.~\ref{topk2_sup}. 
It is shown that $\alpha$ controls the strength of editing and $k$ controls the degree of disentanglement. Higher $\alpha$ leads to more intense editing, and lower $k$ leads to more disentangled manipulation.
We provide more qualitative results of image manipulation with truncated features in Fig.~\ref{topk1_sup}, demonstrating the effectiveness of our approach for disentangled image manipulation.

\begin{table}[]

\resizebox{1\linewidth}{!}{
\begin{tabular}{l|ccccc}
\toprule[1.2pt]
    & \textbf{HDAE(U)}  & SPADE~\cite{Park2019SPADE} & CLADE~\cite{Tan2022CLADE} & GroupDNet~\cite{Zhu2020GroupDNet} & Pix2PixHD~\cite{Wang2018Pix2PixHD} \\ \midrule[1.2pt]
FID & \textbf{23.37} & 29.2  & 30.6  & 25.9      & 38.5   \\ 
 \bottomrule[1.2pt]
\end{tabular}
}
\caption{\textbf{Image translation results of HDAE(U), SPADE~\cite{Park2019SPADE}, CLADE~\cite{Tan2022CLADE}, GroupDNet~\cite{Zhu2020GroupDNet}, and Pix2PixHD~\cite{Wang2018Pix2PixHD} on the CelebAMask-HQ~\cite{maskgan} test set.}}
\label{tab_s2i_sup}
\end{table}

\newcommand{\textred}[1]{\color[HTML]{FE0000}{#1}}
\newcommand{\textgreen}[1]{\color[HTML]{00c001}{#1}}
\newcommand{\textblue}[1]{\color[HTML]{00c3fb}{#1}}
\begin{table}[]

\resizebox{1\linewidth}{!}{
\begin{tabular}{l|l|cccc}
\toprule[1.2pt]

\multirow{2}{*}{Dataset}& \multirow{2}{*}{Model}    & \multicolumn{4}{c}{FID($\downarrow$)} \\
                        &                                                       &T=10                    &T=20    &T=50     &T=100 \\ \midrule[1.2pt]
CelebA 64   
                        & DAE$^\ast$                                               &\textbf{12.92}                   &10.18   &7.05     &5.30   \\ 
                        & HDAE-M(E)  &13.19                   &\textbf{9.86}    &\textbf{6.63}     &\textbf{5.13}   \\ \midrule[1.2pt]
FFHQ 128       
                        & DAE$^\ast$                                              &21.24                   &\textbf{17.15}   &\textbf{13.08}     &\textbf{10.93}   \\ 
                       & HDAE-M(E)  &\textbf{20.73}          &17.36            & 14.24             &12.66            \\ \midrule[1.2pt]
Horse 128               
                        & DAE$^\ast$                                              &12.60                   &10.23   &8.57     &\textbf{8.02}   \\ 
                       & HDAE-M(U) &\textbf{11.29}     &\textbf{9.79}   &\textbf{8.39}     &8.22       \\ 

\bottomrule[1.2pt]
\end{tabular}
}

\caption{
\textbf{Unconditional image generation results of DAE~\cite{DAE} and HDAE-M(E) on the CelebA, FFHQ and LSUN Horse.}
$\ast$ denotes results produced by our re-implementation.
}
\label{gen}
\end{table}

\begin{figure}[t]
    \centering
    \includegraphics[width=0.94\linewidth]{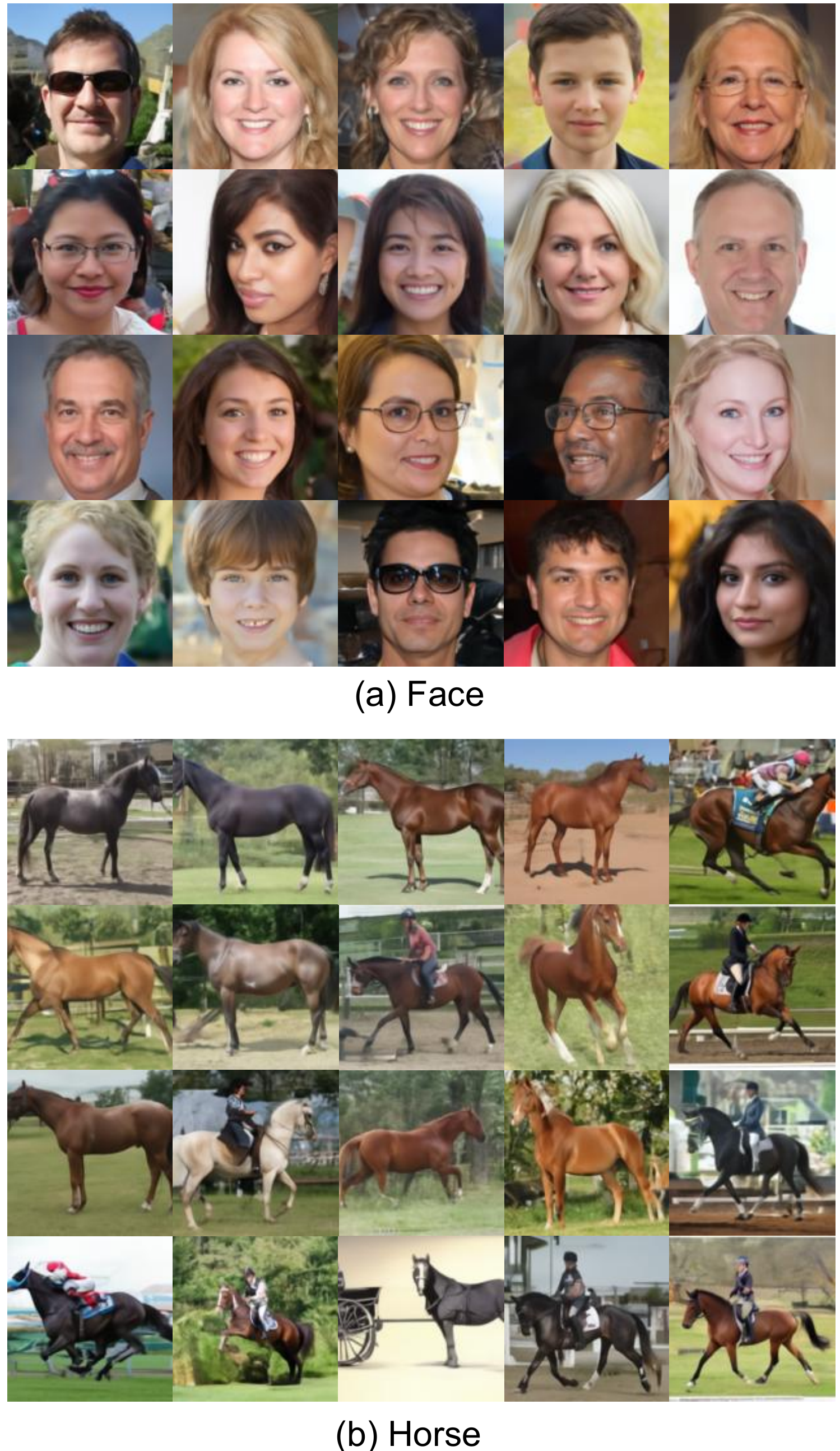}
    \caption{
        \textbf{Unconditional samples (uncurated) from our HDAE-M and latent DDIM trained on FFHQ-128 and LSUN horse-128.}
    }
    \label{fim_sup}
    \vspace{0.1cm}
\end{figure}

\subsection{Interpreting the Hierarchical Latent Space}
\begin{figure*}[!h]
    \centering
    \includegraphics[width=0.94\linewidth]{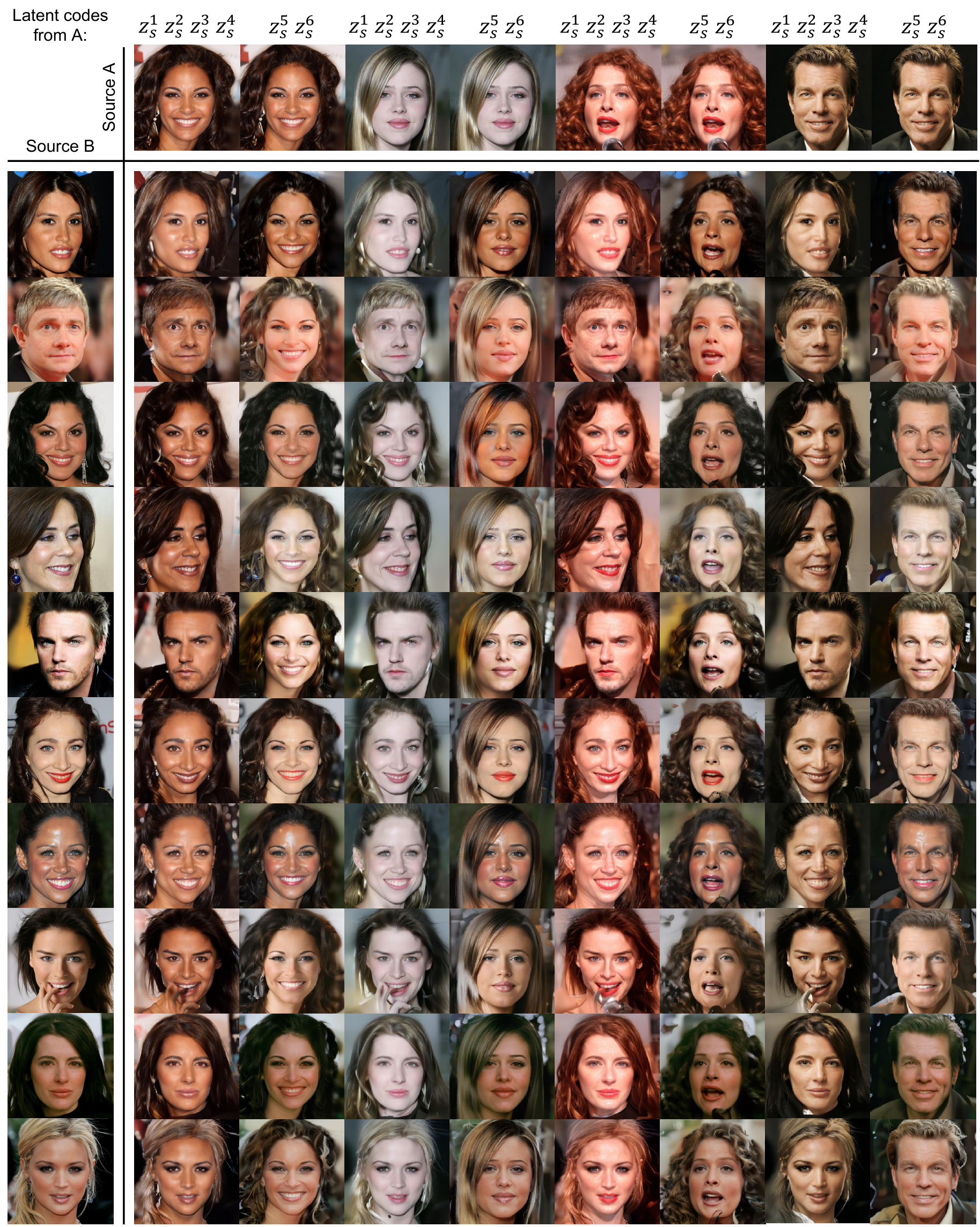}
    \caption{
        \textbf{Style mixing results with hierarchical latent space.}
    }
    \vspace{0.8cm}
    \label{stylemix_sup}
\end{figure*}

\noindent\textbf{Style mixing.}
We show more qualitative results of style mixing in Fig.~\ref{stylemix_sup}.

\begin{figure*}[!h]
    \centering
    \includegraphics[width=0.8\linewidth]{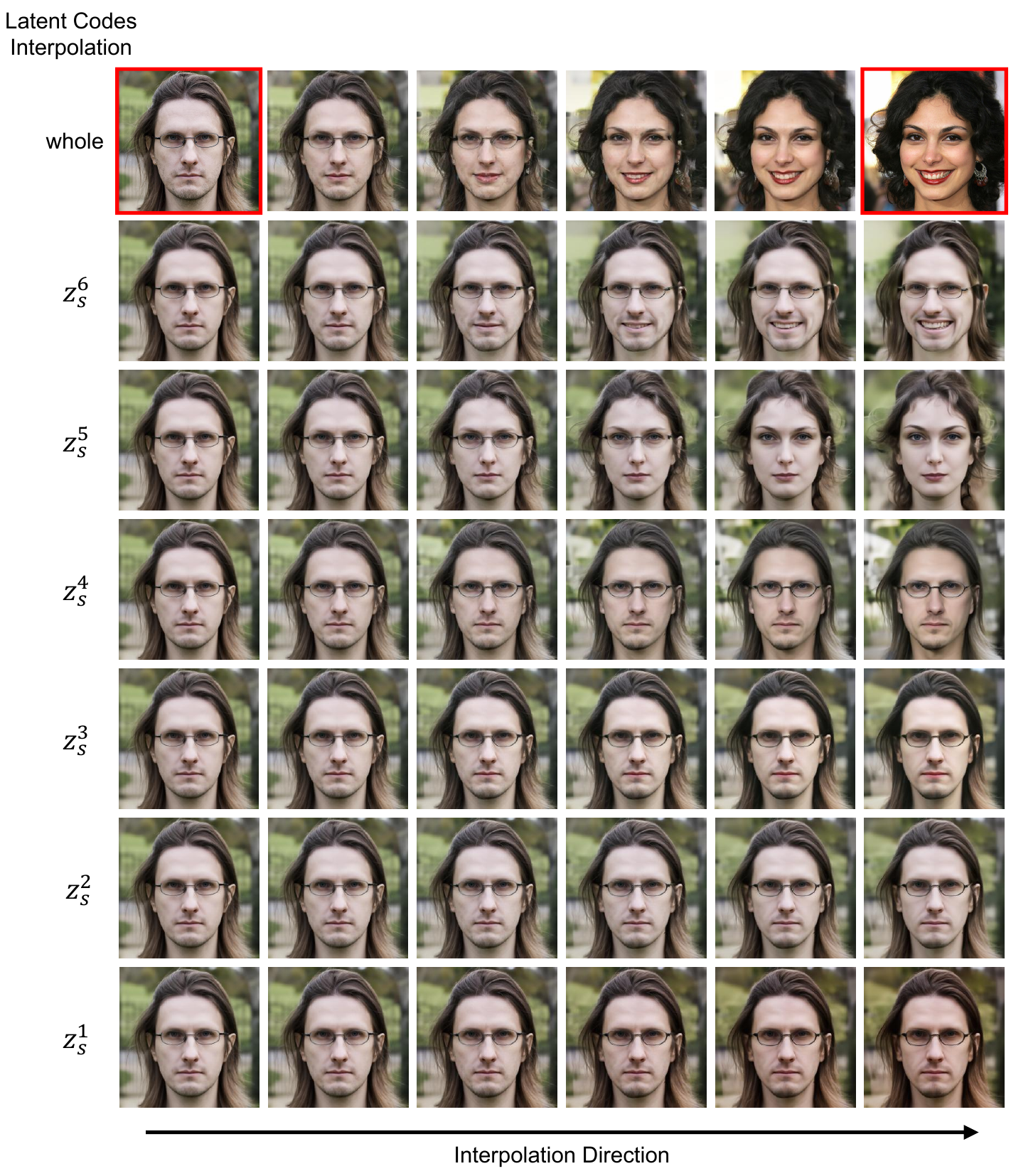}
    \caption{
        \textbf{Image interpolation results with different latent codes.}
    }
    \vspace{2cm}
    \label{interpolate_sup}
\end{figure*}

\noindent\textbf{Image interpolation with different latent codes.}
We show the qualitative results of image interpolation with different latent codes in Fig.~\ref{interpolate_sup}.
It shows a clear hierarchy of the latent space. The low-level features control the spatial details such as background, color, and lighting, and  high-level features control the global and abstract semantic attributes related to image structure such as pose, gender, face shape, and eyeglasses.

\begin{figure*}[!h]
    \centering
    \includegraphics[width=0.95\linewidth]{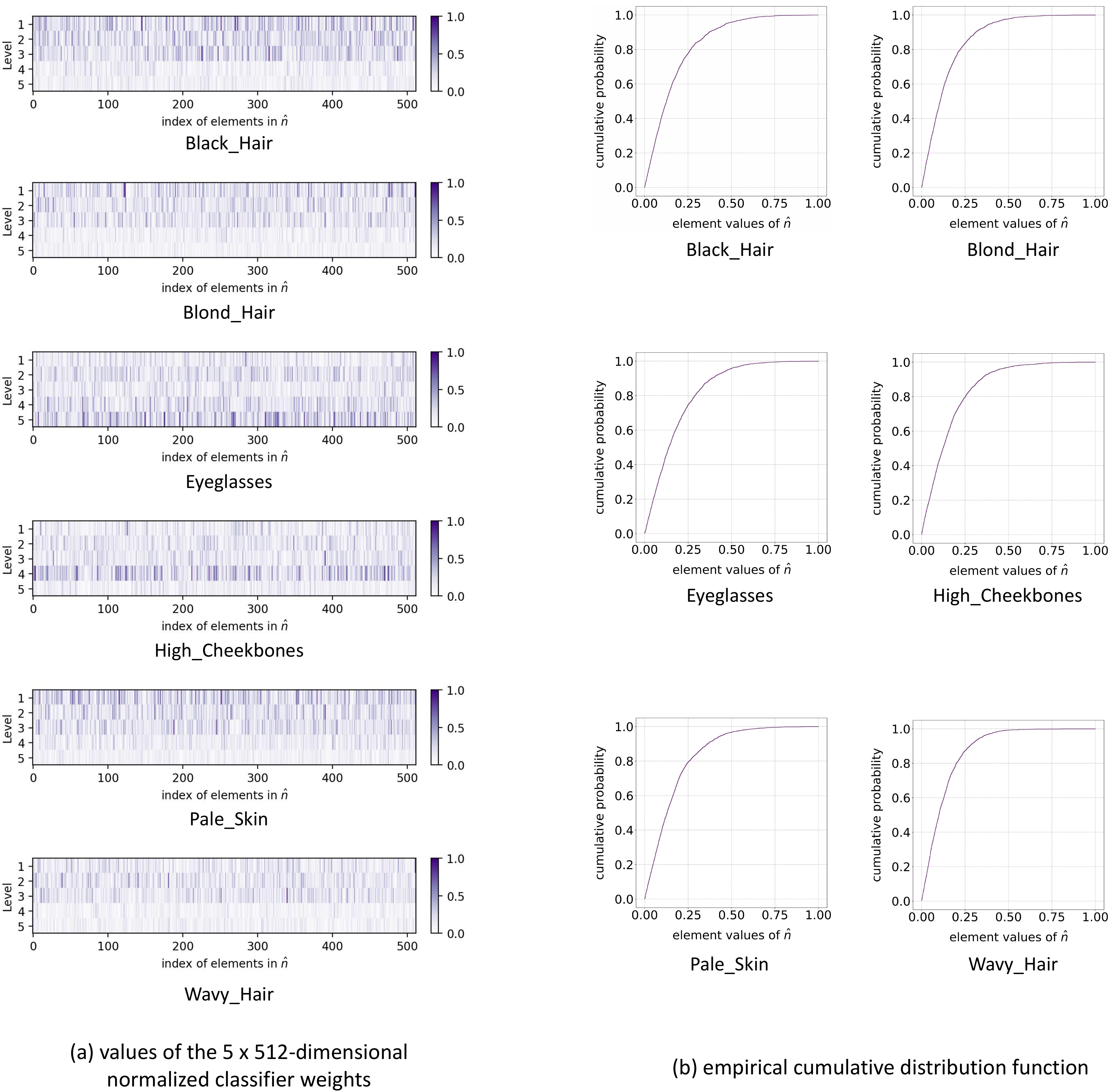}

    \caption{
        \textbf{
            More examples of the values of the $5\times512$-dimensional $\hat{n}$, visualized by levels and the empirical cumulative distribution function of the element values in the normalized classifier weights $\hat{n}$.
        }
    }

    \label{cdf}
\end{figure*}

\noindent\textbf{Visualization of empirical cumulative distribution function and values of the 5 x 512-dimensional normalized classifier weights.}
We show more examples of empirical cumulative distribution function and values of the 5 x 512-dimensional normalized classifier weights in Fig.~\ref{cdf}.

\subsection{Details of human perceptual experiments.}
\begin{figure*}[!h]
    \vspace{-0.2cm}
    \centering
    \includegraphics[width=1\linewidth]{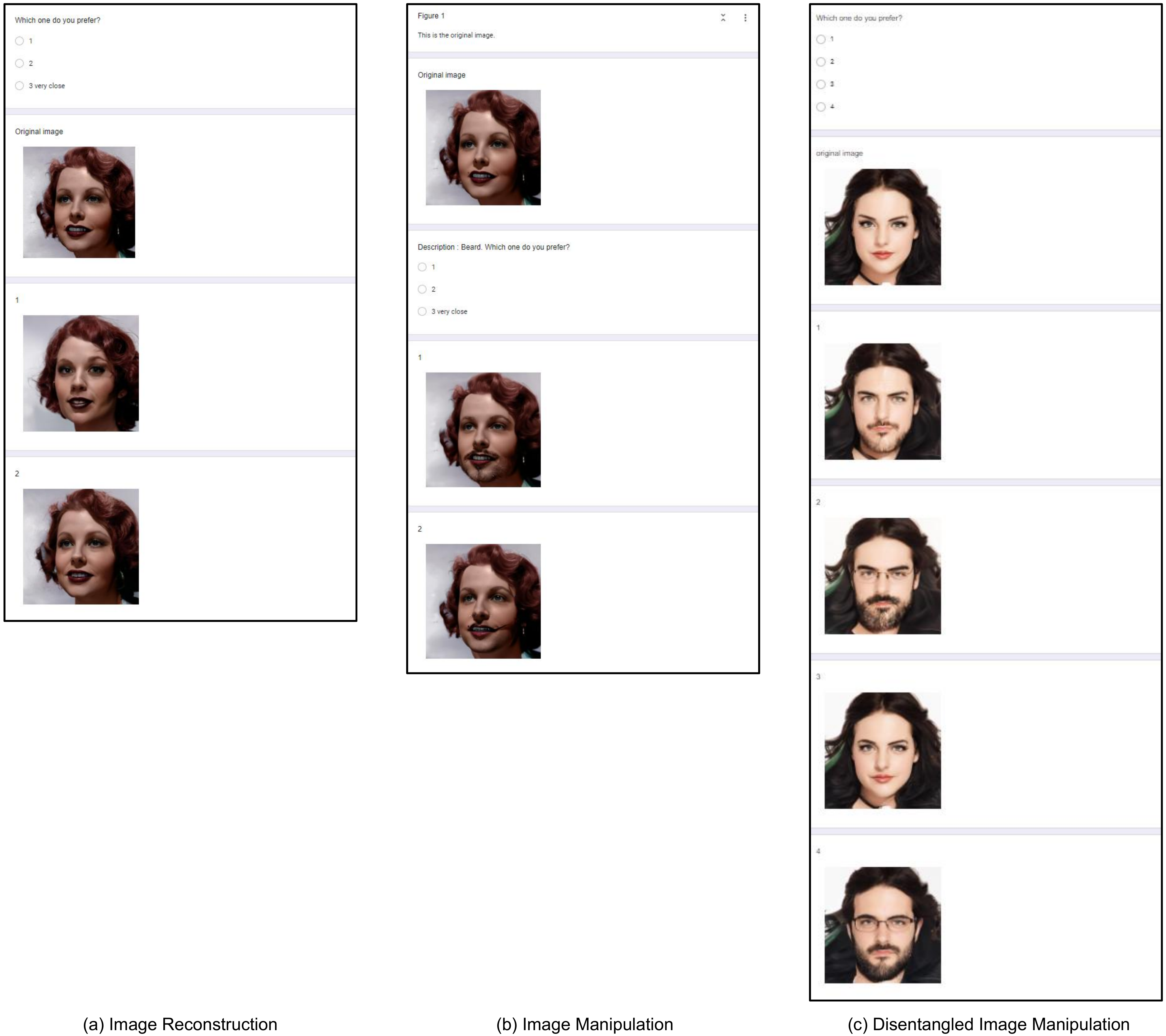}
    \caption{
        \textbf{
            User study examples of image reconstruction, image manipulation and disentangled image manipulation.
        }
    }
    \label{user_study}
\end{figure*}

We collect votes from 15 participants for our human perceptual experiments.
Each participant answers 35 three-choice questions for image manipulation experiment, 35 four-choice questions for disentangled manipulation experiment, and 15 two-choice questions for image reconstruction experiment.
We show some examples of our human perceptual experiments in Fig. ~\ref{user_study}.

\subsection{Semantic Image Synthesis}

\begin{figure*}[!h]
    \centering
    \includegraphics[width=0.9\linewidth]{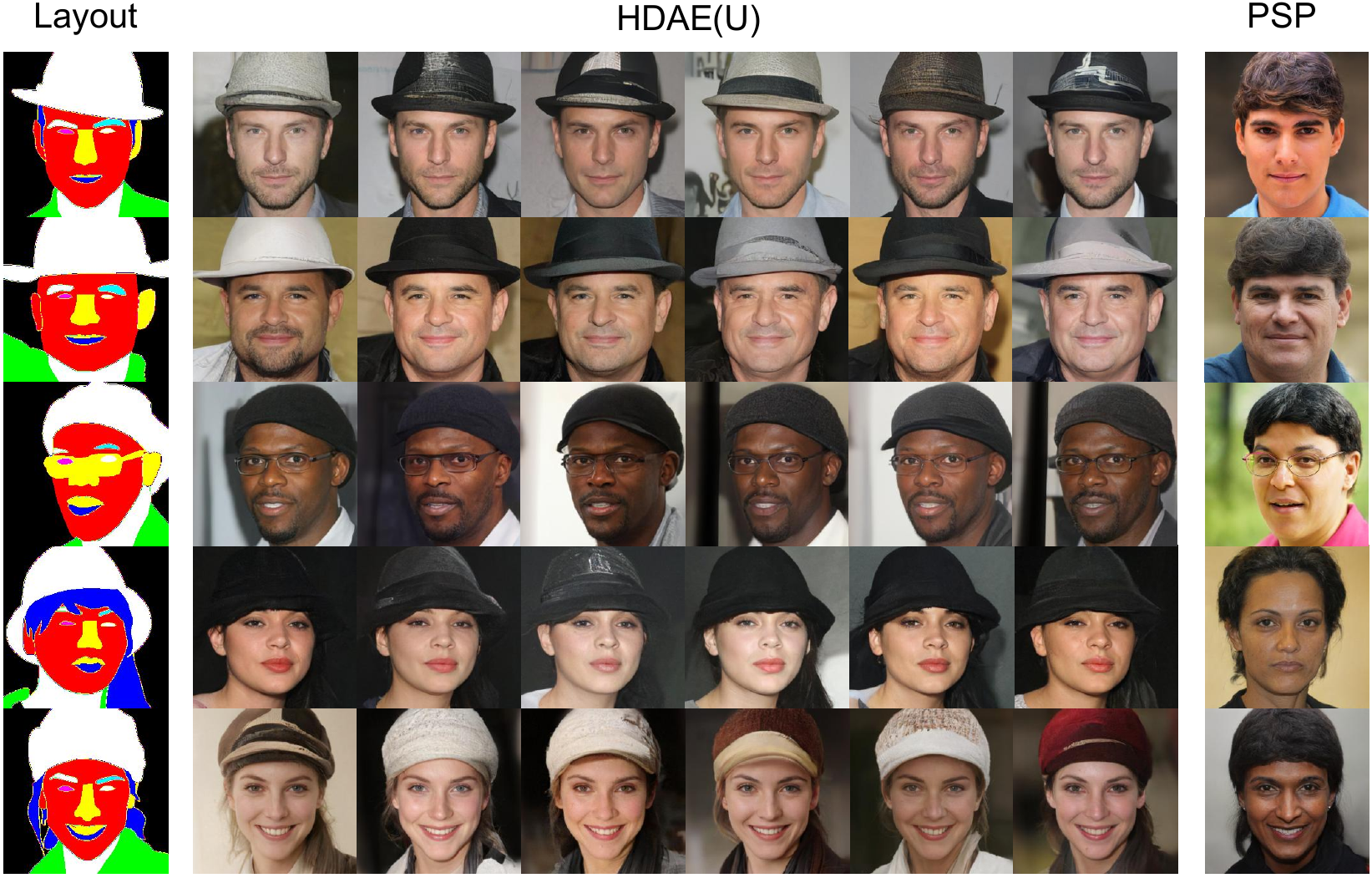}
    \caption{
        \textbf{Semantic image synthesis results between HDAE(U) and PSP~\cite{PSP}}
    }
    \label{s2i_sup}
\end{figure*}

We use HDAE(U) to transform semantic layouts into realistic images.
To fully leverage the semantic information, the semantic label map is injected into the semantic encoder pretrained for semantic layouts to obtain semantic vectors $z_s$.
The stochastic code $x_T$ is a randomly sampled Gaussian noise map.
Our HDAE(U) is trained on the CelebAMask-HQ~\cite{maskgan} dataset with image sizes of $256 \times 256$.
Tab.~\ref{tab_s2i_sup} reports the results of HDAE(U), SPADE~\cite{Park2019SPADE}, CLADE~\cite{Tan2022CLADE}, GroupDNet~\cite{Zhu2020GroupDNet}, and Pix2PixHD~\cite{Wang2018Pix2PixHD}.
As shown in Fig.~\ref{s2i_sup}, HDAE(U) can produce a superior performance on fidelity and learned correspondence without any special design for this task.
By sampling different Gaussian noise maps $x_T$, the model can produce diverse high-quality images with the same layout.

\subsection{Unconditional Image Generation}

\noindent\textbf{Methods.}
We conduct the unconditional image generation experiments on CelebA~\cite{celeba}, FFHQ~\cite{Stylegan}, and LSUN Horse~\cite{Lsun}.
Since the dimension of our hierarchical semantic vectors is higher than the dimension of the semantic vector of DAE, it is more difficult for HDAE than DAE to predict the latent semantic vectors with a latent DDIM.
So we use a linear layer to map our hierarchical semantic vectors from HDAE encoder into a 512-dimensional vector. 
And we use this 512-dimensional vector as the condition of the diffusion U-Net network.
We denote the model as HDAE-M.
After tuning (or training from scratch) our HDAE-M few epochs, we train a latent DDIM to generate the semantic latent code from random noise.


\noindent\textbf{Experiments.}
We compute FID scores between $50,000$ randomly sampled real images from the dataset and our $50,000$ generated images.
Tab.~\ref{gen} reports our experiments on CelebA with image size $64\times64$, FFHQ of image size $128\times128$, and LSUN Horse of image size $128\times128$.
Our approach performs better than DAE on two out of three datasets.
We show more qualitative results of unconditional image generation in Fig.~\ref{gen}.



\label{sec:appendix}

 

\end{document}